\documentclass[11pt]{article}

\usepackage[preprint]{acl}

\usepackage{times}
\usepackage{latexsym}
\usepackage[T1]{fontenc}
\usepackage[utf8]{inputenc}
\usepackage{microtype}
\usepackage{inconsolata}
\usepackage{amsmath}
\usepackage{amssymb}
\usepackage{array}
\usepackage{algorithm}
\usepackage{algpseudocode}
\usepackage{booktabs}
\usepackage{float}
\usepackage{multirow}
\usepackage{graphicx}
\usepackage{xspace}
\usepackage{enumitem}
\usepackage{placeins}
\usepackage{tcolorbox}
\tcbuselibrary{skins, breakable, listings}
\usepackage[table]{xcolor}
\usepackage{colortbl}
\usepackage{arydshln}
\usepackage{fontawesome5}
\usepackage{cleveref}

\definecolor{bestshade}{HTML}{D6E6F5}
\definecolor{secondshade}{HTML}{ECF3FA}
\definecolor{iconfull}{HTML}{2A9D8F}
\definecolor{iconempty}{HTML}{C44536}
\definecolor{iconpartial}{HTML}{9AA5B1}
\definecolor{iconlearn}{HTML}{1F4E79}
\definecolor{iconrand}{HTML}{E76F51}
\definecolor{iconfrozen}{HTML}{4A90A4}
\definecolor{deltacolor}{HTML}{2A7D5C}
\definecolor{deltadown}{HTML}{C44536}
\definecolor{goldicon}{HTML}{D4AF37}
\definecolor{silvericon}{HTML}{9AA5B1}
\newcommand{\drop}[1]{\textcolor{deltacolor}{$\boldsymbol{#1}$}}
\newcommand{\dropdown}[1]{\textcolor{deltadown}{$\boldsymbol{#1}$}}
\newcommand{\champ}{\textcolor{goldicon}{\faMedal}}
\newcommand{\runnerup}{\textcolor{silvericon}{\faMedal}}

\newcommand{\iconYes}{\textcolor{iconfull}{\faCheckCircle}}
\newcommand{\iconNo}{\textcolor{iconempty}{\faTimesCircle}}
\newcommand{\iconHalf}{\textcolor{iconpartial}{\faAdjust}}
\newcommand{\iconLearn}{\textcolor{iconlearn}{\faGraduationCap}}
\newcommand{\iconRand}{\textcolor{iconrand}{\faDice}}
\newcommand{\iconFrozen}{\textcolor{iconfrozen}{\faSnowflake}}
\newcommand{\iconFwd}{\textcolor{iconpartial}{\faPlayCircle}}

\definecolor{phasecolor}{HTML}{1F4E79}
\newcommand{\phaseline}[1]{\textcolor{phasecolor}{\textit{\textbf{#1}}}}

\newcommand{\roles}{\mathcal{P}}          

\newcommand{\cmark}{\textcolor{green!55!black}{\ensuremath{\boldsymbol{\checkmark}}}}
\newcommand{\pmark}{\textcolor{orange!85!black}{\ensuremath{\boldsymbol{\circ}}}}
\newcommand{\xmark}{\textcolor{red!70!black}{\ensuremath{\boldsymbol{\times}}}}

\crefname{section}{\S}{\S\S}
\Crefname{section}{\S}{\S\S}
\crefname{subsection}{\S}{\S\S}
\Crefname{subsection}{\S}{\S\S}
\crefname{equation}{Eq.}{Eqs.}
\Crefname{equation}{Eq.}{Eqs.}
\creflabelformat{equation}{#2#1#3}
\newcommand{\appref}[1]{Appendix~\ref{#1}}

\newcommand{\sero}{\textsc{Sero}\xspace}
\newcommand{\naturalplan}{\textsc{NaturalPlan}\xspace}
\newcommand{\tablebench}{\textsc{TableBench}\xspace}
\newcommand{\olympiadbench}{\textsc{OlympiadBench}\xspace}

\title{Roles with Rails: Contract-Preserving Role Evolution \\ in Multi-Agent Structured Reasoning}

\author{
\normalfont\mdseries
Ling-Yue Ge\textsuperscript{1,2}, Lan-Zhe Guo\textsuperscript{1,2*}\\
\textsuperscript{1}National Key Laboratory for Novel Software Technology, Nanjing University, Nanjing, China\\
\textsuperscript{2}School of Intelligence Science and Technology, Nanjing University, Suzhou, China\\
\textsuperscript{*}Corresponding author: \texttt{guolz@lamda.nju.edu.cn}
}

\begin{document}
\maketitle

\begin{abstract}
Role-based LLM multi-agent systems need adaptive role pools, yet adapting such systems is not merely a matter of prompt optimization: roles often carry structural obligations, including capability coverage, message compatibility, validation, final-answer aggregation, and parser-compatible output protocols. Existing systems either fix the role inventory and lose adaptivity, or allow unconstrained generation to induce role drift, removing structurally necessary roles and breaking answer contracts. We formulate this as \emph{contract-preserving role evolution}, requiring every committed edit to preserve five structural contracts (capability, communication, validation, aggregation, output protocol). We instantiate this formulation in \sero, a \textbf{S}elf-\textbf{E}volving \textbf{R}ole \textbf{O}rchestration framework that evolves a typed role-card pool through credit-guided retrieval, a credit-ranked communication DAG with a protected terminal aggregator and conditional validator repair, and a contextual-bandit controller whose LLM-proposed edits are committed only when they preserve the contracts and improve task score. Experiments on real-world reasoning benchmarks across three LLM backbones confirm the value of contract-preserving role evolution.
\end{abstract}

\section{Introduction}
\label{sec:introduction}

\begin{figure*}[tbh]
\centering
\includegraphics[width=.9\textwidth]{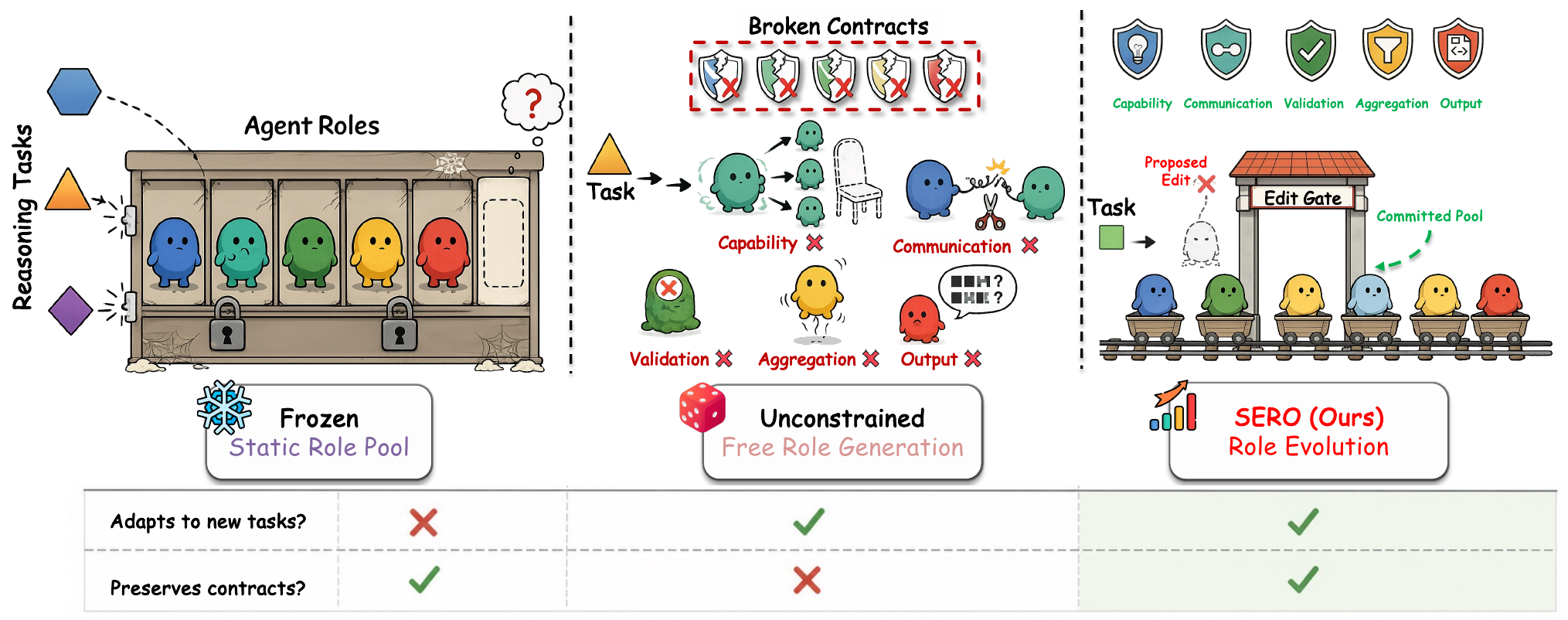}
\caption{Three role-pool paradigms for multi-agent LLM reasoning. \textbf{Left}: a frozen pool preserves the five contracts (capability, communication, validation, aggregation, output protocol) but admits no new roles. \textbf{Middle}: unconstrained editing breaks one or more. \textbf{Right}: \sero gates each edit, committing only those preserving all five.}
\label{fig:overview}
\end{figure*}

Role-based LLM multi-agent systems decompose reasoning across specialized agents such as planners, solvers, critics, validators, and aggregators, supporting task decomposition across specialist agents~\citep{hong2024metagpt}, multi-agent debate~\citep{du2024improving}, and graph-structured conversational coordination~\citep{Wu2023AutoGenEN}. Many deployed systems use a fixed role set or fixed interaction template and topology per task family. These fixed decompositions are stable precisely because their roles preserve capability, communication, validation, aggregation, and output-format contracts. Yet these same contracts make the role set fundamentally hard to adapt when the task distribution shifts.

Existing attempts to make role-based agents adaptive break this tension from only one side. Fixed-topology and pruning methods learn task-conditioned communication graphs or remove redundant agents and edges \citep{zhuge2024gptswarm,zhang2025g,wang2025agentdropout,zhang2025cut}, preserving task decomposition structure but cannot introduce new role capabilities. Dynamic role, prompt, and behavior adaptation methods rewrite roles or topologies, assign debate participants, or co-evolve agent behavior \citep{wang2026metagen,zhang2026dynamic,chen2025multi}, but unconstrained role updates could be unstable, removing useful roles, duplicating capabilities, or breaking answer protocols.  \textbf{Fig.~\ref{fig:overview}} contrasts these regimes: frozen pools preserve contracts but add no roles; unconstrained editing adds roles but breaks contracts, leaving the challenge of doing both.

We formulate this problem as \emph{contract-preserving role evolution}, in which a multi-agent system edits a reusable role pool while preserving the structural contracts required for coordination, checking, and final-answer production.  An edit is valid only when it preserves five structural contracts. The \emph{capability} contract keeps required expertise covered, preventing evolution from pruning the last role for a task family. The \emph{communication} contract keeps intermediate states exchangeable under declared protocols, preventing incompatible or disconnected handoffs. The \emph{validation} contract keeps error-detection or repair capacity reachable, preventing unchecked drafts from bypassing critique. The \emph{aggregation} contract keeps a protected terminal role responsible for the final decision, preventing competing or missing final answers. The \emph{{output-protocol}} contract keeps the final response compatible with the benchmark parser or application interface. Contract-preserving role evolution is therefore guarded editing over typed role cards, not unconstrained prompt search.

We realize contract-preserving role evolution as \sero (\textbf{S}elf-\textbf{E}volving \textbf{R}ole \textbf{O}rchestration, Right of \textbf{Fig.~\ref{fig:overview}}). Each agent is a typed role card carrying capability metadata, a communication protocol, a role type, and a protection flag, so contract checks operate on structured fields rather than raw prompts. Role-pool edits are framed as $\textsc{Add}$, $\textsc{Remove}$, or $\textsc{Noop}$ decisions over a contextual-bandit controller, masked by the five contracts and committed only when they improve the task score under the same inference operator. At inference, \sero retrieves a small active team and routes it through a credit-ranked DAG terminating at a protected aggregator with optional validator repair, so each candidate pool is tested by the same executable contract check used during commitment. Our contributions are:
\begin{itemize}
  \item \textbf{Formalism.} We define \emph{contract-preserving role evolution} through five contracts and cast it as guarded editing over typed role cards.
  \item \textbf{Framework.} We propose \sero, a framework that admits an edit when it preserves five contracts and persists it only when it improves the task score under the same inference operator.
  \item \textbf{Mechanism.} \sero introduces typed role cards as contract unit, multi-scale credit for retrieval and removal, and a contract-masked, score-gated contextual-bandit controller.
  \item \textbf{Evidence.} Extensive experiments on three structured-reasoning benchmarks across three LLM backbones establish \sero's superiority over single-agent and multi-agent baselines.
\end{itemize}

\section{Related Work}
\label{sec:related}

\paragraph{Topology and Pruning.}
LLM multi-agent systems decompose complex reasoning tasks across specialized agents and have increasingly been built around role-playing societies, software-engineering teams, debate-style frameworks, and tool-coordination platforms \citep{li2023camel,hong2024metagpt,du2024improving,Wu2023AutoGenEN,shen2023hugginggpt}. Once a role inventory has been established, two adaptation strategies dominate existing designs. Topology methods learn task-conditioned communication graphs over a fixed inventory \citep{zhuge2024gptswarm,zhang2025g,leong2025amas,zhao2026sc,fan2026todycomm,chen2026goagent}. Selection and pruning methods drop redundant agents or edges and assign per-agent credit during execution using difference rewards, leave-one-out effects, or Shapley-style estimates \citep{zhang2025cut,wang2025agentdropout,li2025adaptive,wolpert2001optimal,shapley201617,nagpal2025leveraging,NEURIPS2025_80164880}. These methods optimize how a role inventory communicates and which subset acts, yet leave the long-term composition of the inventory largely underexplored.
\begin{table*}[tbh]
\centering
\footnotesize
\setlength{\tabcolsep}{5pt}
\renewcommand{\arraystretch}{1.0}
\newcommand{\lightrowrule}{\arrayrulecolor{black!30}\hline\arrayrulecolor{black}}
\resizebox{.94\textwidth}{!}{
\begin{tabular}{@{}p{0.48\textwidth} cccccc@{}}
\arrayrulecolor{black}\toprule
& Inventory & Topology & Edit & Cross-Task & Role & Verification \\
System Family & Adaptation & Adaptation & Ops & Persistence & Typing & Loop \\
\midrule
CAMEL \citep{li2023camel} / MetaGPT \citep{hong2024metagpt} / AutoGen \citep{Wu2023AutoGenEN}
  & \xmark & \pmark & \xmark & \xmark & \pmark & \pmark \\ \lightrowrule
DyLAN \citep{liu2023dynamic} / MAD \citep{du2024improving} / DynamicRole \citep{zhang2026dynamic}
  & \pmark & \pmark & \pmark & \xmark & \xmark & \cmark \\ \lightrowrule
GPTSwarm \citep{zhuge2024gptswarm} / G-Designer \citep{zhang2025g} / AMAS \citep{leong2025amas}
  & \xmark & \cmark & \xmark & \pmark & \xmark & \xmark \\ \lightrowrule
ARG-Designer \citep{li2026assemble} / GoAgent \citep{chen2026goagent} / SC-MAS \citep{zhao2026sc} / TodyComm \citep{fan2026todycomm}
  & \pmark & \cmark & \pmark & \xmark & \xmark & \xmark \\ \lightrowrule
AgentPrune \citep{zhang2025cut} / AgentDropout \citep{wang2025agentdropout} / AGP \citep{li2025adaptive}
  & \pmark & \cmark & \pmark & \pmark & \xmark & \xmark \\ \lightrowrule
MetaGen \citep{wang2026metagen}
  & \cmark & \cmark & \pmark & \xmark & \xmark & \pmark \\ \lightrowrule
MAE \citep{chen2025multi}
  & \xmark & \xmark & \xmark & \pmark & \xmark & \pmark \\
\arrayrulecolor{black}\midrule
\textbf{\sero} (Ours, this paper)
  & \cmark & \cmark & \cmark & \cmark & \cmark & \cmark \\
\bottomrule
\end{tabular}
}
\arrayrulecolor{black}
\caption{\small LLM multi-agent systems on six design dimensions: \emph{Inventory Adaptation} (role set adapts without redeployment), \emph{Topology Adaptation} (per-task communication graph), \emph{Edit Ops} (Add/Remove; \pmark{} also covers selection, pruning, or prompt rewrite without persistent role-card edits), \emph{Cross-Task Persistence} (changes persist), \emph{Role Typing} (typed objects vs prompts), \emph{Verification Loop} (output-acceptance signal). \cmark{}~supports, \pmark{}~partial, \xmark{}~absent.}
\label{tab:related_comparison}
\end{table*}

\paragraph{Dynamic Role and Prompt Adaptation.}
Several mechanisms adapt a multi-agent system's roles, prompts, or composition during a run. MetaGen evolves roles and topologies jointly \citep{wang2026metagen}, and online role assignment picks debate participants and models for role slots dynamically during an interaction \citep{zhang2026dynamic,liu2023dynamic}. Query-conditioned methods generate topologies and choose agents from an extensible pool \citep{li2026assemble}, and online prompt optimization updates agent prompts from feedback signals \citep{xia2026hivemind}. Cooperative evolution updates role behavior through iterativeself-play in a fixed template \citep{chen2025multi}. These methods improve adaptivity, but unconstrained updates to roles, prompts, or compositions can be unstable when they persist beyond a single decision, removing useful roles, duplicating capabilities, or breaking validation, aggregation, and output protocols. \sero instead treats role-set adaptation as \emph{contract-preserving role evolution}, committing role-pool edits only when they preserve structural contracts and measurably improve the task score, and contrasts with the closest systems along six axes in \textbf{Table~\ref{tab:related_comparison}}.

\section{Methodology: \sero}
\label{sec:method}

\paragraph{Problem Formulation.}
We consider LLM multi-agent reasoning in which a persistent pool of typed roles collaborates to answer tasks drawn from a training distribution $p_{\mathrm{train}}$. Let $\roles$ denote this role pool and $F$ an inference operator that, given a task $x$, assembles a task-conditioned active subset $A\subseteq\roles$ and produces an answer $\hat y=F(x;\roles)$ scored by a benchmark scorer $s_x(\cdot)$. We seek
\begin{equation}
  \roles^{\star}=\arg\max_{\roles}\;\mathbb{E}_{x\sim p_{\mathrm{train}}}\!\bigl[s_x(F(x;\roles))\bigr],
  \label{eq:objective}
\end{equation}
subject to the requirement that every committed edit to $\roles$ preserves the five structural contracts of \emph{contract-preserving role evolution} (capability, communication, validation, aggregation, output protocol), so the pool remains executable under $F$. Optimizing \cref{eq:objective} by mutating prompts on per-task signals from $s_x$ violates these contracts and yields pools that fail on non-trivial subsets despite high single-task scores. Throughout, $e(\cdot)$ denotes a shared sentence encoder applied uniformly to task text, role-card prompts, and inference messages.

\begin{figure*}[tbh]
\centering
\includegraphics[width=.95\textwidth,keepaspectratio]{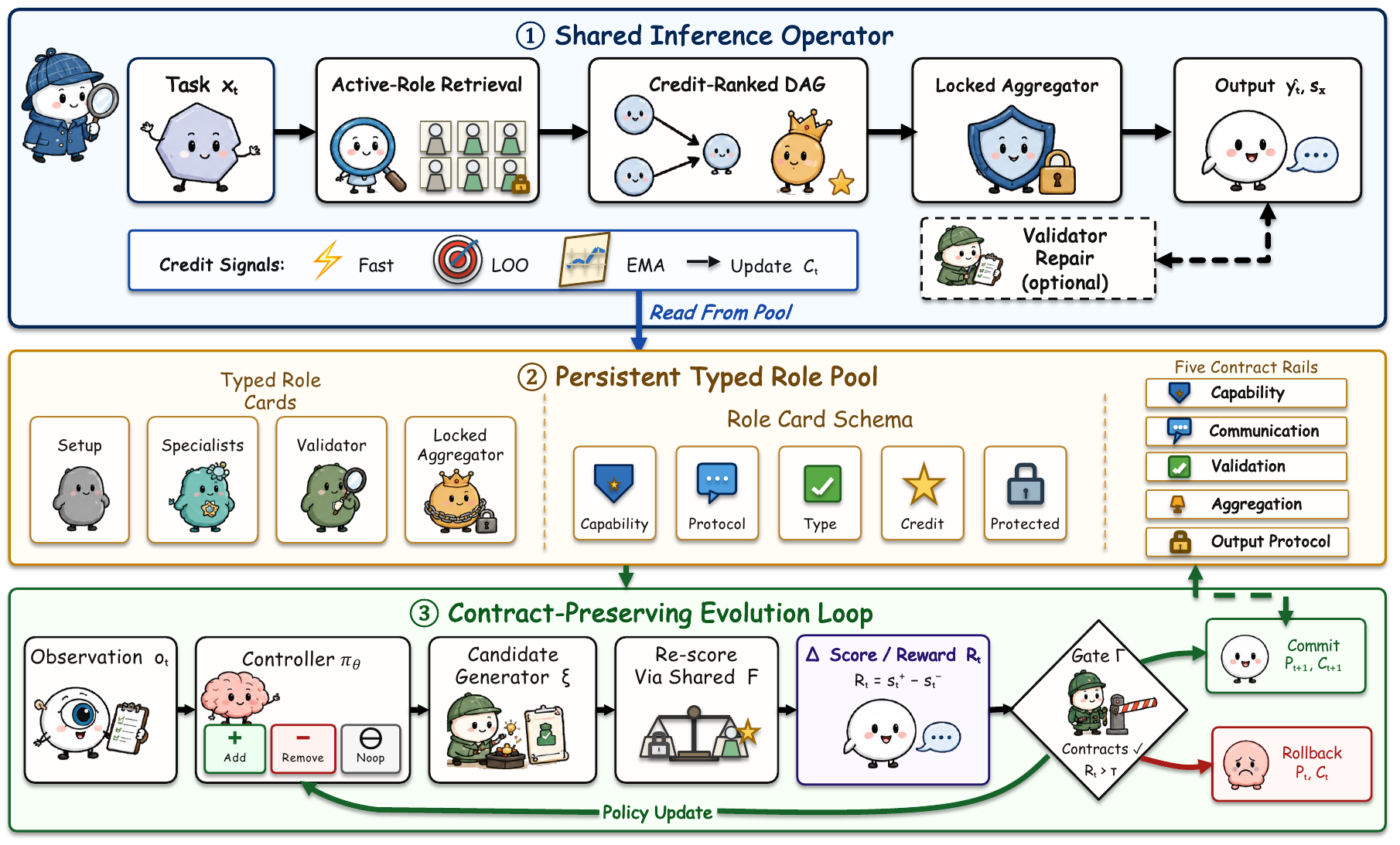}
\caption{Framework overview of \sero. \textbf{\textcircled{1}} Shared inference operator $F$ retrieves an active team, runs a credit-ranked DAG ending at a locked aggregator with optional validator repair, updating credit via fast, leave-one-out, and EMA signals. \textbf{\textcircled{2}} A typed role pool stores cards (capability, protocol, type, credit, protection), with a locked validator and aggregator anchoring the five contract rails. \textbf{\textcircled{3}} Controller $\pi_\theta$ proposes an $\textsc{Add}/\textsc{Remove}/\textsc{Noop}$ edit, generator $\mathcal{E}$ builds a candidate pool, $F$ re-scores it for reward $R_t$, and $\Gamma$ commits or rolls back.}
\label{fig:method_pipeline}
\end{figure*}

\paragraph{Overview.}
\sero (\textbf{Fig.~\ref{fig:method_pipeline}}) approaches \cref{eq:objective} as a contextual-bandit~\cite{langford2007epoch} problem in which edits to $\roles_t$ are proposed freely across training steps and committed only when they both \emph{satisfy the five contracts} and \emph{improve the task score} under the same inference operator. Each component of \sero implements one face of this rule. Role cards make contracts explicit on each editable unit (\Cref{sec:role_cards}). Credit estimates identify which roles can be safely retrieved, kept, or removed over time (\Cref{sec:credit}). The inference operator $F$ acts as the executable contract test, retrieving a query-specific team, routing it through a credit-ranked DAG, and producing a protected answer with optional validator repair (\Cref{sec:inference}). A guarded evolution loop binds proposals to contract-defined action masks and a score-gated commitment rule (\Cref{sec:guarded_evolution}).

\subsection{Role Cards}
\label{sec:role_cards}

Our editable unit is a typed role card $r_i$ whose fields map directly to the contract dimensions,
\begin{equation}
  r_i = \langle n_i,p_i,g_i,f_i,\ell_i,T_i,\nu_i,b_i\rangle.
  \label{eq:role_card}
\end{equation}
Tags $g_i$ and family $f_i$ encode capability, $\ell_i$ specifies a communication protocol, $\nu_i\in\{\mathsf{Spec},\mathsf{Setup},\mathsf{Val},\mathsf{Agg}\}$ is a role type ($\mathsf{Setup}$ covers early task framing such as contract parsing), and $b_i\in\{0,1\}$ indicates whether a role is structurally necessary, with $b_i=1$ for the terminal aggregator and required validators. The remaining fields $n_i$, $p_i$, $T_i$ carry name, system prompt, and decoding temperature. Every commit must preserve all five contracts. \textbf{(i)~Capability.}~The pool retains at least one role per required capability family. \textbf{(ii)~Communication.}~The active team forms a DAG consistent with each role's protocol $\ell_i$. \textbf{(iii)~Validation.}~A validator slot is reserved when post-aggregation repair is enabled. \textbf{(iv)~Aggregation.}~A protected terminal aggregator is appended outside the specialist budget. \textbf{(v)~Output protocol.}~$r_{\mathrm{agg}}$'s protocol enforces the benchmark-required answer format.

\subsection{Credit Estimation}
\label{sec:credit}

Credit is the evidence used to decide which roles can be safely retrieved, kept, or removed. Three signals provide it at complementary scales. Fast credit gives low-cost within-run attribution, precise leave-one-out credit calibrates whether a role was actually necessary on a task, and historical EMA credit prevents single-instance noise from causing unstable edits. Let $m_i$ denote the message produced by role $r_i$ in the current inference pass and $\bar e_M=\tfrac{1}{|A_t|}\sum_{r_i\in A_t} e(m_i)$ the consensus embedding of active-role messages. Fast credit is a within-task proxy computable on every step, as
\begin{equation}
  c_i^{\mathrm{fast}}
  =\beta\cos(e(m_i),e_x)
  +(1-\beta)\cos(e(m_i),\bar e_M),
  \label{eq:fast_credit}
\end{equation}
where $e_x=e(x)$, the first term favors task alignment, and the second favors agreement with the active set. Validator fast credit is treated as a structured signal rather than a similarity score, and increases when validator feedback is adopted by a repair pass. Precise credit refines fast credit with periodic leave-one-out evaluation to approximate causal contribution under the current pool,
\begin{equation}
  \hat\phi_i = s_x(F(x;\roles_t,C_t))
  - s_x(F(x;\roles_t\setminus\{r_i\},C_t)),
  \label{eq:loo_credit}
\end{equation}
and historical credit tracks longer-run value via the EMA update~\citep{morales2024exponential}, which can be formulated as:
\begin{equation}
  \bar c_i \;\mathrel{\mathop:}=\; (1-\mu)\,\bar c_i + \mu\,\hat\phi_i.
  \label{eq:ema_credit}
\end{equation}
New roles start with $\bar c_i=0$ and conservative update counts so exploration is possible without letting unevaluated roles dominate retrieval.

\subsection{Inference Procedure}
\label{sec:inference}

Inference is the executable contract test of the current role pool. The operator $F$ retrieves a compact team, routes information through protocol-compatible roles, produces a protected answer, and applies validator repair where needed. The same $F$ is applied at evaluation and during training, so the candidate-vs-current score difference reflects the behavior induced by the evaluated edit.

\paragraph{Active-Role Retrieval.}
Candidate roles are ranked by a convex combination of semantic relevance and historical credit while always including the protected aggregator card $r_{\mathrm{agg}}$ identified by $\nu_i=\mathsf{Agg}$, which can be formulated as:
\begin{equation}
  \rho(r_i,x)
  =\alpha\cos(e(p_i),e_x)
  +(1-\alpha)\,\mathrm{norm}(\bar c_i),
  \label{eq:retrieval}
\end{equation}
where $e_x=e(x)$ and $\mathrm{norm}(\cdot)$ rescales EMA credit to $[0,1]$ across the current pool. When validator passes are enabled, separate retrieval budgets are allocated for specialists and validators so critique does not compete with upstream reasoning roles, and $r_{\mathrm{agg}}$ is appended after retrieval so the terminal contract is preserved. The retrieved set $A_t$ becomes the vertex set for message scheduling.

\paragraph{Credit-Ranked DAG Construction.}
Message scheduling uses a stage-aware order that prefers reliable contributors and reserves late positions for validation. Let $A_t$ be the active set and $B_t=A_t\setminus\{r_{\mathrm{agg}}\}$ the non-terminal roles, and define a lexicographic key as follows:
\begin{equation}
  K_i=(\kappa(\nu_i),\,-c_i^{\mathrm{fast}},\,i),\qquad
  r_i\prec r_j \Leftrightarrow K_i<K_j,
  \label{eq:dag_order}
\end{equation}
where $\kappa(\mathsf{Setup})=-1$, $\kappa(\mathsf{Spec})=0$, $\kappa(\mathsf{Val})=1$, and $c_i^{\mathrm{fast}}$ comes from \Cref{eq:fast_credit}. Edges are added greedily from earlier to later roles under fixed in-degree and out-degree caps, and every non-terminal role is additionally connected to $r_{\mathrm{agg}}$ so the terminal aggregator always observes all upstream content even when intermediate coordination is sparse. The induced DAG is partitioned into dependency levels, roles at the same level are invoked in parallel, and each role receives the original task, upstream messages, and its role-card protocol. \Cref{alg:credit_ranked_dag} gives the full construction used in the experiments.

\paragraph{Aggregation and Validator Repair.}
The terminal stage routes all messages to the protected aggregator $r_{\mathrm{agg}}$ and produces the answer scored by $s_x(\cdot)$. An optional validator intervenes on contract-risky drafts. If the validator flags an issue, $r_{\mathrm{agg}}$ is re-invoked with the draft and validator feedback~\citep{madaan2023self}, and the resulting output $\hat y_t$ feeds both the controller's reward signal and the credit updates in \Cref{eq:fast_credit,eq:ema_credit}.

\subsection{Guarded Evolution}
\label{sec:guarded_evolution}

We separate \emph{proposing} from \emph{committing} via three components, a controller policy $\pi_\theta$ selecting $a_t=(\omega_t,z_t)$ with $\omega_t\in\mathcal{O}=\{\textsc{Add},\textsc{Remove},\textsc{Noop}\}$ and $z_t$ a target role for $\textsc{Remove}$, a candidate generator $\mathcal{E}$ realizing $\omega_t$ as a typed candidate pool $(\roles'_t,C'_t)$ under structural constraints, and a commitment rule $\Gamma$ deciding persistence from the candidate-vs-current score change
\begin{equation}
  R_t = s_x\bigl(F(x_t;\roles'_t,C'_t)\bigr) - s_x\bigl(F(x_t;\roles_t,C_t)\bigr),
  \label{eq:reward}
\end{equation}
with $R_t=0$ for $\textsc{Noop}$, invalid edits, or unchanged candidates. The training objective is $J(\theta)=\mathbb{E}_{\pi_\theta}\!\left[\sum_{t=1}^{T} R_t\right]$ under an action mask induced by structural invariants.

\begin{table*}[t!]
\centering
\footnotesize
\setlength{\tabcolsep}{5.5pt}
\renewcommand{\arraystretch}{1.18}
\resizebox{.87\textwidth}{!}{%
\begin{tabular}{@{}ll cc c c c c@{}}
\toprule
\multirow{2}{*}{Backbone} & \multirow{2}{*}{Method}
  & \multicolumn{2}{c}{\textsc{NaturalPlan}}
  & \multirow{2}{*}{\textsc{Olympiad}}
  & \multirow{2}{*}{\textsc{Table}}
  & \multirow{2}{*}{Avg.}
  & \multirow{2}{*}{$\Delta$} \\
\cmidrule(lr){3-4}
 &  & Partial & Exact &  &  &  &  \\
\midrule
\multirow[t]{7}{*}{GPT-4o-mini}
  & CoT                       & 50.26$_{0.52}$ & 22.11$_{0.19}$ & 37.71$_{0.51}$ & 47.24$_{0.26}$ & 39.33$_{0.10}$ & \drop{+6.30} \\
  & SC@3                        & 50.61$_{0.09}$ & 22.15$_{0.55}$ & \cellcolor{secondshade}\underline{39.61}$_{1.09}$ & 47.99$_{1.15}$ & 40.09$_{0.16}$ & \drop{+5.54} \\
  & Static DAG MAS            & 55.99$_{1.17}$ & 21.11$_{0.78}$ & 38.46$_{0.33}$ & 52.30$_{1.16}$ & 41.97$_{0.70}$ & \drop{+3.66} \\
  & Workflow                  & \cellcolor{secondshade}\underline{57.16}$_{0.12}$ & 24.78$_{1.16}$ & 34.88$_{1.58}$ & 49.08$_{0.85}$ & 41.48$_{0.51}$ & \drop{+4.15} \\
  & Static Role Orchestration~\runnerup & 56.91$_{0.98}$ & \cellcolor{secondshade}\underline{28.15}$_{0.90}$ & 38.97$_{0.21}$ & \cellcolor{secondshade}\underline{52.43}$_{1.31}$ & \cellcolor{secondshade}\underline{44.12}$_{0.13}$ & \drop{+1.51} \\
  & Random Role Evolution     & 55.11$_{3.08}$ & 25.26$_{3.00}$ & 38.54$_{3.34}$ & 51.76$_{0.45}$ & 42.67$_{2.31}$ & \drop{+2.96} \\
  & \sero~(Ours)~\champ & \cellcolor{bestshade}\textbf{57.82}$_{0.69}$ & \cellcolor{bestshade}\textbf{30.37}$_{0.93}$ & \cellcolor{bestshade}\textbf{40.27}$_{0.74}$ & \cellcolor{bestshade}\textbf{54.06}$_{0.59}$ & \cellcolor{bestshade}\textbf{45.63}$_{0.49}$ & --- \\
\midrule
\multirow[t]{7}{*}{Gemini-2.5-flash-lite}
  & CoT                       & 59.34$_{0.18}$ & 30.96$_{0.13}$ & 40.93$_{0.16}$ & 59.00$_{0.27}$ & 47.56$_{0.05}$ & \drop{+18.97} \\
  & SC@3                        & 59.29$_{0.14}$ & 30.48$_{0.51}$ & 40.92$_{0.67}$ & 58.50$_{0.69}$ & 47.30$_{0.08}$ & \drop{+19.23} \\
  & Static DAG MAS~\runnerup   & \cellcolor{secondshade}\underline{77.68}$_{0.66}$ & \cellcolor{secondshade}\underline{54.70}$_{0.89}$ & 51.02$_{0.65}$ & 60.68$_{0.13}$ & \cellcolor{secondshade}\underline{61.02}$_{0.52}$ & \drop{+5.51} \\
  & Workflow                  & 72.57$_{0.11}$ & 51.15$_{0.34}$ & \cellcolor{secondshade}\underline{60.74}$_{0.55}$ & 57.91$_{0.51}$ & 60.59$_{0.19}$ & \drop{+5.94} \\
  & Static Role Orchestration & 70.04$_{0.73}$ & 43.93$_{0.45}$ & 52.97$_{10.68}$ & \cellcolor{secondshade}\underline{61.60}$_{0.26}$ & 57.13$_{2.84}$ & \drop{+9.40} \\
  & Random Role Evolution     & 63.41$_{7.24}$ & 40.70$_{4.00}$ & 55.97$_{4.62}$ & 60.64$_{1.06}$ & 55.18$_{2.10}$ & \drop{+11.35} \\
  & \sero~(Ours)~\champ & \cellcolor{bestshade}\textbf{80.71}$_{0.62}$ & \cellcolor{bestshade}\textbf{56.78}$_{0.48}$ & \cellcolor{bestshade}\textbf{65.15}$_{1.00}$ & \cellcolor{bestshade}\textbf{63.48}$_{1.58}$ & \cellcolor{bestshade}\textbf{66.53}$_{0.74}$ & --- \\
\midrule
\multirow[t]{7}{*}{Qwen3-8b}
  & CoT                       & 46.10$_{0.11}$ & 13.81$_{0.17}$ & 35.57$_{0.69}$ & 29.44$_{0.07}$ & 31.23$_{0.14}$ & \drop{+8.61} \\
  & SC@3                        & 46.19$_{0.10}$ & 13.81$_{0.06}$ & 36.60$_{0.94}$ & 29.27$_{0.13}$ & 31.47$_{0.22}$ & \drop{+8.37} \\
  & Static DAG MAS~\champ     & \cellcolor{bestshade}\textbf{54.90}$_{0.55}$ & \cellcolor{bestshade}\textbf{20.30}$_{1.07}$ & 42.90$_{0.68}$ & 44.05$_{1.03}$ & \cellcolor{bestshade}\textbf{40.54}$_{0.38}$ & \dropdown{-0.70} \\
  & Workflow                  & \cellcolor{secondshade}\underline{50.42}$_{0.31}$ & \cellcolor{secondshade}\underline{15.78}$_{0.22}$ & 40.07$_{1.62}$ & 44.43$_{0.53}$ & 37.68$_{0.60}$ & \drop{+2.16} \\
  & Static Role Orchestration & 47.92$_{0.20}$ & 13.59$_{0.23}$ & \cellcolor{secondshade}\underline{47.89}$_{1.22}$ & 47.95$_{0.89}$ & 39.34$_{0.38}$ & \drop{+0.50} \\
  & Random Role Evolution     & 39.28$_{16.95}$ & 10.26$_{8.42}$ & 47.54$_{4.90}$ & \cellcolor{secondshade}\underline{48.32}$_{3.64}$ & 36.35$_{5.08}$ & \drop{+3.49} \\
  & \sero~(Ours)~\runnerup & 46.44$_{0.58}$ & 14.96$_{1.28}$ & \cellcolor{bestshade}\textbf{48.60}$_{0.69}$ & \cellcolor{bestshade}\textbf{49.37}$_{0.22}$ & \cellcolor{secondshade}\underline{39.84}$_{0.35}$ & --- \\
\bottomrule
\end{tabular}%
}
\caption{Main results over three runs. \textbf{Best} and \underline{second-best} per column; $\Delta$ is \sero's gain on the row average. \champ~/~\runnerup~mark the top-1 and top-2 method per backbone by average score. Full results are provided in \textbf{Table~\ref{tab:app_seed_all}}.}
\label{tab:main_results}
\vspace{-4pt}
\end{table*}

\paragraph{Controller.}
The controller proposes edits, while the contract checker defines which proposals are admissible and the score gate decides which admissible proposals become persistent. A pre-decision inference pass with $F$ produces a provisional answer $\hat y_t$ and an active set $A_t\subseteq\roles_t$, which are summarized into the observation
\begin{equation}
  o_t=\bigl[e(x_t,\hat y_t),\;\bar e_{A_t},\;\psi(C_t)\bigr],
  \label{eq:observation}
\end{equation}
where $e(x_t,\hat y_t)$ is the embedding of the task concatenated with answer, $\bar e_{A_t}=\tfrac{1}{|A_t|}\sum_{r_i\in A_t} e(p_i)$ summarizes the assembled team and $\psi(C_t)\in\mathbb{R}^{5}$ collects pool-level credit statistics,
\begin{equation}
  \psi(C_t)=\bigl(\bar c,\;\sigma_c,\;c_{\min},\;c_{\max},\;\bar{\hat\phi}^{\mathrm{recent}}\bigr).
  \label{eq:credit_state_summary}
\end{equation}
Here, $\bar c$, $\sigma_c$, $c_{\min}$, and $c_{\max}$ denote the mean, standard deviation, minimum, and maximum historical credit over the current pool, and $\bar{\hat\phi}^{\mathrm{recent}}$ denotes the mean recent leave-one-out credit.
The first two slots condition edits on observed task-level behavior and on the semantic composition of the assembled team. Writing $\chi_t\in\{\textsc{Warmup},\textsc{Main}\}$ for the training phase, the credit slot is retained but masked to zero during $\textsc{Warmup}$ so early exploration does not depend on stale credit memory. A shared encoder maps $o_t$ to $h_t$, an operation head produces a distribution over $\mathcal{O}$, and a per-role head scores admissible removal targets using each role's prompt embedding and local credit features. For $\textsc{Remove}$, with per-role score $\xi_i(o_t)$ and admissible target set $\mathcal{V}_t$, the target is sampled from the masked conditional
\begin{equation}
  p_\theta(z_t{=}r_i\mid\omega_t{=}\textsc{Remove},o_t)
  = \frac{\exp\!\bigl(\xi_i(o_t)\bigr)}
         {\sum_{j\in\mathcal{V}_t}\exp\!\bigl(\xi_j(o_t)\bigr)}.
\label{eq:target_sampling}
\end{equation}
$\textsc{Add}$ has no controller-level target and creates a new card using the highest-EMA role selected inside $\mathcal{E}$ as its anchor. The mask enforces the invariants from \Cref{sec:role_cards}, including phase-dependent constraints via $\chi_t$, protection flags $b_i$, and minimum coverage requirements. We train $\pi_\theta$ with REINFORCE \citep{williams1992simple} using batch-normalized rewards and an EMA baseline.

\paragraph{Candidate Generator $\mathcal{E}$.}
Given the controller's action $a_t=(\omega_t,z_t)$, $\mathcal{E}$ produces a schema-conforming, contract-checked candidate pool before scoring. For $\textsc{Add}$, $\mathcal{E}$ uses a role editor anchored on the highest-EMA role to generate a typed card $r_{\text{new}}$ from task context, the current pool, credit summaries, and coverage diagnostics, and rejects candidates that violate the schema, over-concentrate dominant capability families, or near-duplicate existing prompts. For $\textsc{Remove}$, $\mathcal{E}$ proposes deleting a role only when removal preserves all contracts, and for $\textsc{Noop}$ it returns the current pool unchanged. The output is a candidate state $(\roles'_t,C'_t)$ evaluated through $F$ and passed to $\Gamma$.

\paragraph{Commitment Rule $\Gamma$.}
A candidate edit is committed only if it both satisfies the five contracts, enforced by the action mask and the constructor checks above, and improves the task score over the unedited pool. The score-side test uses $R_t$ in \Cref{eq:reward} with phase-dependent thresholds controlled by $\chi_t$, kept conservative to filter single-task noise and prevent gradual drift from many small commits. If $\Gamma$ rejects a real edit, $(\roles_t,C_t)$ is restored exactly, while the controller is still updated using the observed $R_t$. Training loop in \Cref{alg:training_step} .

\section{Experiments}
\label{sec:experiments}

\paragraph{Benchmarks and Metrics.}
We evaluate on three benchmark families chosen to stress different structural demands and answer contracts.  \naturalplan \citep{zheng2024natural} mixes trip planning, calendar scheduling, and meeting planning, and we report both partial and exact task accuracy.  For \tablebench \citep{wu2025tablebench}, we use the non-visual subset covering fact checking, numerical reasoning, and data analysis over text-only tables, scored as normalized single-line answer accuracy.  For \olympiadbench \citep{he2024olympiadbench}, we use the text-only English open-ended question subset covering competition mathematics and physics, scored with the official judge and string-matching fallback.  Held-out split statistics and metric definitions are reported in \appref{app:benchmark_details}.

\begin{figure*}[tbh]
\centering
\includegraphics[width=.92\textwidth]{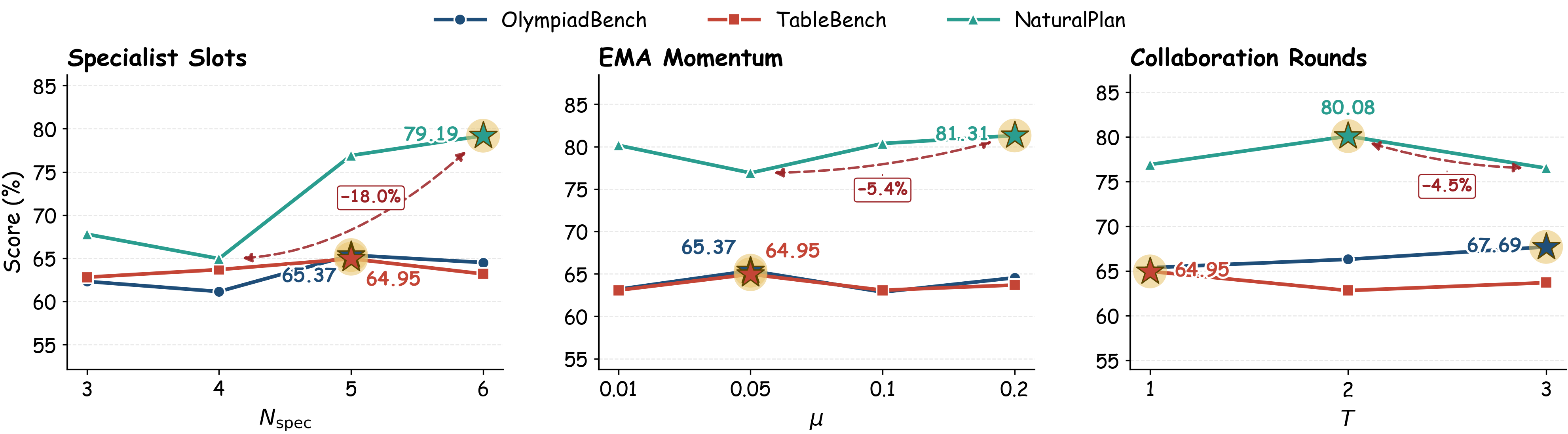}
\caption{Hyper-parameter sensitivity of \sero under one-at-a-time sweeps. Scores are task accuracy; the \naturalplan curve uses partial accuracy. Stars mark the best score per benchmark.}
\label{fig:hp_sensitivity}
\end{figure*}

\begin{table*}[tbh]
\centering
\footnotesize
\setlength{\tabcolsep}{4.5pt}
\renewcommand{\arraystretch}{1.28}
\resizebox{.92\textwidth}{!}{%
\begin{tabular}{lcccccccccccc}
\toprule
\multirow{2}{*}{Method} & \multirow{2}{*}{Evol.} & \multirow{2}{*}{Credit} & \multirow{2}{*}{Prot./Val.} & \multirow{2}{*}{Ctrl.}
 & \multicolumn{2}{c}{\olympiadbench}
 & \multicolumn{2}{c}{\tablebench}
 & \multicolumn{2}{c}{\naturalplan~(Partial)}
 & \multicolumn{2}{c}{\naturalplan~(Exact)} \\
\cmidrule(lr){6-7}\cmidrule(lr){8-9}\cmidrule(lr){10-11}\cmidrule(lr){12-13}
 &  &  &  &  & Score & $\Delta$ & Score & $\Delta$ & Score & $\Delta$ & Score & $\Delta$ \\
\midrule
Static Role Orchestration    & \iconNo     & \iconHalf & \iconHalf & \iconNo  & 64.42 & \drop{-0.95}  & 61.31 & \drop{-3.64} & 70.03 & \drop{-11.28} & 44.00 & \drop{-13.33} \\
Random Role Evolution        & \iconRand   & \iconHalf & \iconHalf & \iconNo  & 50.87 & \drop{-14.50} & 61.31 & \drop{-3.64} & 70.92 & \drop{-10.39} & 44.67 & \drop{-12.66} \\
Static DAG MAS               & \iconNo     & \iconNo   & \iconHalf & \iconNo  & 50.28 & \drop{-15.09} & 60.80 & \drop{-4.15} & 76.91 & \drop{-4.40}  & 53.67 & \drop{-3.66}  \\
\midrule
\emph{w/o} Credit            & \iconLearn  & \iconNo   & \iconYes  & \iconYes & 63.67 & \drop{-1.70}  & 63.58 & \drop{-1.37} & 71.29 & \drop{-10.02} & 44.11 & \drop{-13.22} \\
\emph{w/o} Role Evolution    & \iconFrozen & \iconYes  & \iconYes  & \iconYes & 65.07 & \drop{-0.30}  & 63.44 & \drop{-1.51} & 80.15 & \drop{-1.16}  & 56.22 & \drop{-1.11}  \\
\emph{w/o} Protection        & \iconLearn  & \iconYes  & \iconNo   & \iconYes & 64.35 & \drop{-1.02}  & 63.32 & \drop{-1.63} & 80.04 & \drop{-1.27}  & 56.11 & \drop{-1.22}  \\
\emph{w/o} Controller Reward & \iconFwd    & \iconYes  & \iconYes  & \iconNo  & 63.75 & \drop{-1.62}  & 62.81 & \drop{-2.14} & 79.83 & \drop{-1.48}  & 55.89 & \drop{-1.44}  \\
\midrule
\rowcolor{bestshade}
\sero~(Ours)                 & \iconLearn  & \iconYes  & \iconYes  & \iconYes & \textbf{65.37} & --- & \textbf{64.95} & --- & \textbf{81.31} & --- & \textbf{57.33} & --- \\
\bottomrule
\end{tabular}%
}
\caption{Component ablation of \sero. Configurations: \iconLearn~learned, \iconFrozen~frozen pool, \iconRand~random, \iconFwd~forward-only, \iconYes~full / enabled, \iconHalf~default / partial, \iconNo~none / disabled. $\Delta$ is the absolute drop relative to the full method.}
\label{tab:ablations}
\end{table*}

\paragraph{Baselines.}
We compare \sero against six baselines spanning single-agent prompting, fixed workflows, static multi-agent graphs, frozen-pool orchestration, and random role evolution.  \emph{CoT} \citep{wei2022chain} is the single-call baseline, and \emph{SC@3} \citep{wang2022self} uses self-consistency with \(k=3\).  \emph{Workflow} is a benchmark-specific hand-written linear decomposition. \emph{Static DAG MAS} is a benchmark-specific hand-designed non-linear agent graph.  \emph{Static Role Orchestration} keeps \sero's seed pool and the same retrieval-and-DAG inference pipeline but freezes the role pool. \emph{Random Role Evolution} keeps the same edit space and role-card editor as \sero but replaces learned edit selection with uniform random actions. Taken together, these comparisons test whether any gains come mainly from stronger prompting, hand-designed decomposition, generic multi-agent structure, or specifically from learned contract-preserving role evolution. Exact prompts and graph definitions are reported in \appref{app:baselines}.

\paragraph{Implementation Details.}
We report results for GPT-4o-mini \citep{openai2024gpt4omini}, Gemini-2.5-flash-lite \citep{kilpatrick2025gemini25flashlite}, and Qwen3-8b \citep{yang2025qwen3technicalreport}. Within each comparison block, the backbone is frozen; task text, role prompts, and inference messages are embedded by a frozen Jina Embeddings v2 sentence encoder (small, English) \citep{günther2023jina}; and only a lightweight factorized MLP controller for role editing is trainable. Thus, observed differences primarily reflect coordination and adaptation rather than model scale. The controller architecture and implementation are detailed in \appref{app:controller_details}. Trainable variants share the same benchmark-specific training partition and are evaluated on the same held-out split as non-trainable baselines. Split statistics and dataset details are provided in \appref{app:dataset_statistics}.
\subsection{Main Results}
\label{sec:main_results}

\textbf{Table~\ref{tab:main_results}} reports averages over three seeds, with seed-level breakdowns in \appref{app:per_seed}. \sero is the top method on every metric for GPT-4o-mini and Gemini-2.5-flash-lite, lifting the average score by $+1.51$ and $+5.51$ over the strongest baseline respectively, and remains best on \olympiadbench and \tablebench under Qwen3-8b. These results show that credit-guided role evolution yields structural gains beyond what single-agent prompting, hand-written workflows, or frozen multi-agent graphs can supply, with the strongest lift on backbones that can exploit deeper multi-agent coordination. Qwen3-8b on \naturalplan is the exception, where Static DAG MAS retains the lead and marks the scope limit when a smaller open model meets strictly format-constrained planning.

\begin{figure*}[t]
\centering
\includegraphics[width=.89\textwidth]{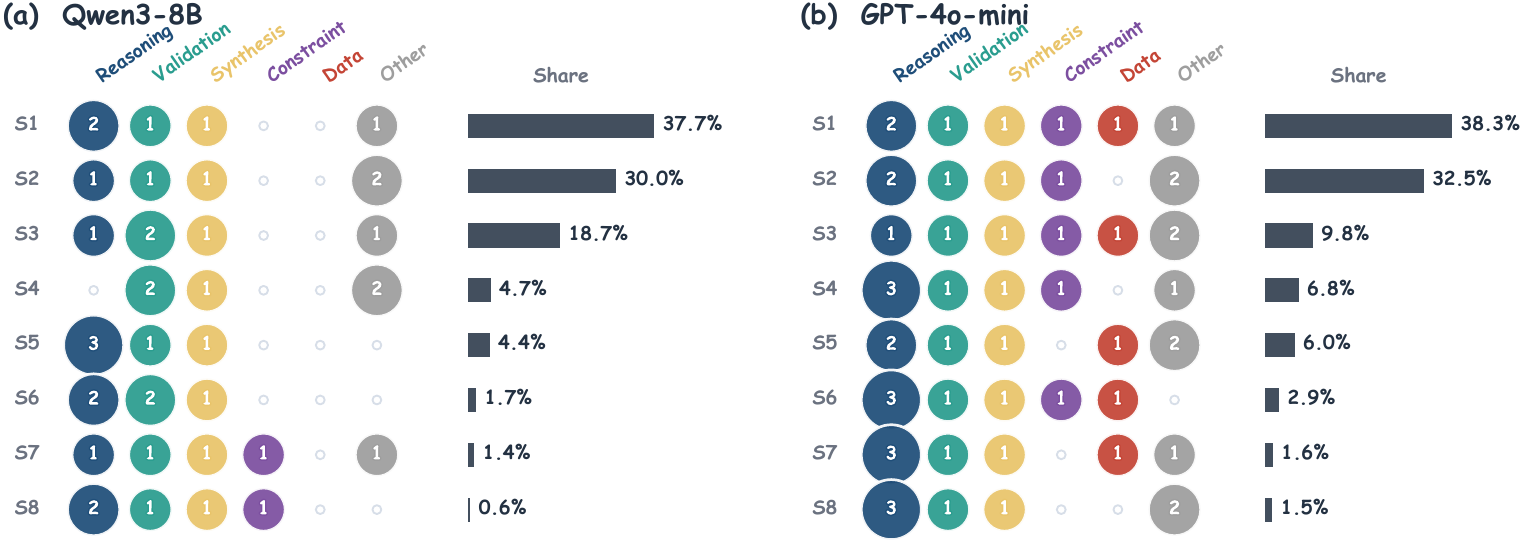}
\caption{Dominant family-level active-set signatures of \sero on \olympiadbench. Each row is a frequent multiset of role families; the bar reports its share among evaluation instances.}
\label{fig:active_set_signatures}
\end{figure*}

\subsection{Framework Analysis}
\label{sec:framework_analysis}

\paragraph{Hyperparameter Sensitivity.}
\label{sec:hp_sensitivity}
\textbf{Fig.~\ref{fig:hp_sensitivity}} reports one-at-a-time sweeps of the specialist budget $N_{\mathrm{spec}}$, credit-EMA momentum $\mu$, and collaboration rounds $T$. \sero stays competitive across the swept range of every axis, though the optimum is benchmark dependent, with math and table benchmarks preferring more diverse specialist pools and lighter smoothing while \naturalplan tolerates the opposite. The gains therefore do not hinge on a single point but track the per-benchmark trade-off between specialist coverage and format discipline.

\paragraph{Ablation Study.}
\label{sec:ablation}
\textbf{Table~\ref{tab:ablations}} isolates each \sero component. The full method tops every metric, removing credit yields the largest drop ($-13.22$ on NP-E and $-10.02$ on NP-P), identifying credit as dominant since it jointly governs edit decisions and role orchestration. Protection, controller reward, and learned evolution add smaller, consistent lifts, while the three non-\sero baselines fall below every \emph{w/o} ablation, indicating gains come from coupling learned role edits with credit-based feedback, rather than static structures or unguided role churn.

\paragraph{Active-Set Signatures.}
\label{sec:active_set_signatures}
\textbf{Fig.~\ref{fig:active_set_signatures}} groups each evaluation instance by the family-level multiset of its active roles. A few dominant signatures capture most routing decisions on both backbones, and each signature keeps a stable Reasoning + Validation + Synthesis core while varying the specialist families around it. This reveals that \sero reuses a small library of coordination templates whose specialist mix is conditioned on task type rather than building a per-instance graph, consistent with gains tracking structured reuse over activation breadth.

\paragraph{Mechanism Analysis.}
\label{sec:mechanism_analyses}
The dominant signatures in \textbf{Fig.~\ref{fig:active_set_signatures}} establish that evaluation-time routing reuses a small set of stable templates rather than constructing per-instance idiosyncratic graphs. \textbf{Table~\ref{tab:compute_overhead}} adds the cost-accuracy view, where \sero matches Static DAG MAS in inference call count and adds only a modest token increase while improving the cross-backbone average score by $+2.83$ absolute, indicating that the gains come from learning reusable, benchmark-dependent coordination templates rather than from invoking more agents. Role-pool lifecycle statistics and routing breakdowns in \appref{app:role_lifecycle_analysis} and \appref{app:active_set_diversity}.

\begin{table}[tbh]
\centering
\footnotesize
\setlength{\tabcolsep}{6pt}
\renewcommand{\arraystretch}{1.12}
\begin{tabular}{lccc}
\toprule
System & Calls/inst. & Tokens/inst. & Avg. \\
\midrule
CoT                          & 1.00 & 1211.88 & 39.37 \\
SC                           & 3.00 & 3634.50 & 39.62 \\
Workflow                     & 4.66 & 5376.96 & 46.58 \\
Static DAG MAS               & 6.31 & 7234.75 & \cellcolor{secondshade}\underline{47.84} \\
\sero~(Ours)                 & 6.31 & 7485.41 & \cellcolor{bestshade}\textbf{50.67} \\
\bottomrule
\end{tabular}
\caption{Per-instance computational overhead and mean accuracy.  Avg.\ averages each method's per-backbone score in \textbf{Table~\ref{tab:main_results}} across the three backbones.}
\label{tab:compute_overhead}
\end{table}

\begin{figure}[tbh]
\centering
\includegraphics[width=\columnwidth]{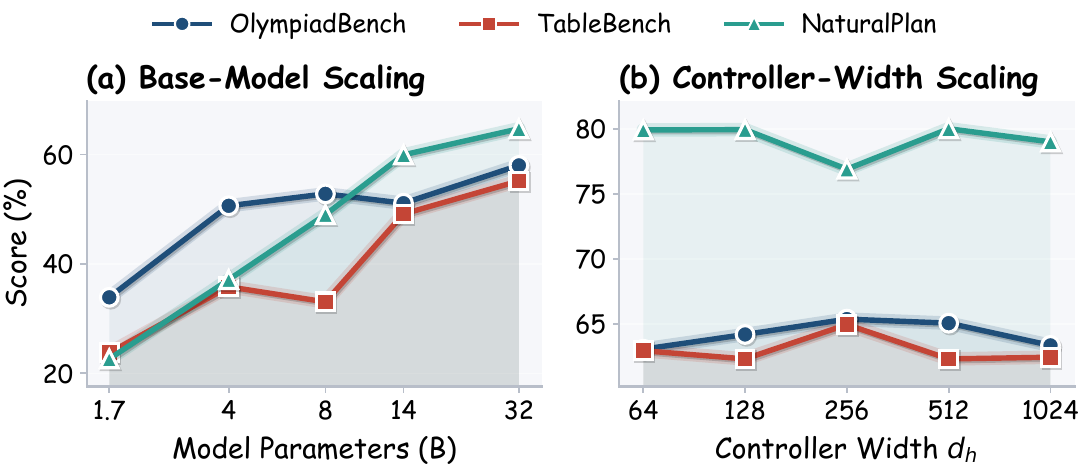}
\caption{Scaling behavior of \sero. \textbf{(a)} Base-model scaling across Qwen3 sizes. \textbf{(b)} Controller-width scaling across hidden width $d_h$ on Gemini-2.5-flash-lite. The \naturalplan curve uses partial accuracy.}
\label{fig:scaling}
\end{figure}

\paragraph{Scaling Behavior.}
\label{sec:scaling}
\textbf{Fig.~\ref{fig:scaling}} separates two scaling axes under \sero. Base-model scaling on Qwen3 (left) lifts all three benchmarks as the model grows from $1.7$B to $32$B, so contract-preserving evolution keeps converting backbone capability into task score. Controller-width scaling on Gemini-2.5-flash-lite (right) is by contrast flat across hidden widths $d_h$ from $64$ to $1024$. The gains therefore track backbone capacity rather than controller capacity, and a small contextual-bandit controller already suffices. Exact numbers in \appref{app:scaling}.

\section{Conclusion}
\label{sec:conclusion}

We propose \sero, a framework realizing \emph{contract-preserving role evolution} over a persistent role pool under five structural contracts. \sero recasts role-pool adaptation as guarded editing, in which candidate edits are proposed freely but committed only when they preserve every contract and improve the task objective. Extensive experiments on \naturalplan, \tablebench, and \olympiadbench across three diverse LLM backbones establish \sero's superiority over advanced baselines.

\clearpage
\section*{Limitations}

\sero is currently restricted to text-only benchmarks with automatically scorable outputs, and its seed role pool is hand-designed by capability family. Extending the framework to multimodal reasoning, multilingual settings, or domains with sparse or subjective rewards remains future work, as does a fully end-to-end accounting of training-time cost across larger backbones.
\bibliography{references}

\clearpage
\appendix
\phantomsection
\section*{Appendix}
\label{app:appendix}
\addcontentsline{toc}{section}{Appendix}

\newtcolorbox{rolecardbox}[1]{enhanced, breakable, sharp corners, boxrule=0.75pt, colframe=black!80, colback=blue!3!white, colbacktitle=black!87, coltitle=white, fonttitle=\bfseries\small, title={#1}, borderline west={2pt}{0pt}{blue!55!black}, left=5pt, right=5pt, top=4pt, bottom=4pt, before skip=0.85em, after skip=0.85em
}
\newcommand{\rolecardentry}[2]{%
\par\noindent\textbf{#1.}\ #2\par\vspace{0.45em}}
\newcommand{\rolecardsubtitle}[1]{%
\par\noindent{\bfseries\footnotesize #1}\par\vspace{0.25em}}

\newtcblisting{rolepromptbox}[1]{enhanced, breakable, sharp corners, boxrule=0.7pt, colframe=blue!45!black, colback=blue!2!white, colbacktitle=blue!70!black, coltitle=white, fonttitle=\bfseries\footnotesize, title={#1}, listing only, listing options={basicstyle=\ttfamily\footnotesize, columns=fullflexible, keepspaces=true, breaklines=true, breakatwhitespace=false, showstringspaces=false, upquote=true, literate={→}{{$\rightarrow$}}1 {←}{{$\leftarrow$}}1 {∥}{{$\parallel$}}1 {×}{{$\times$}}1 {—}{{---}}1 {-}{{-\allowbreak}}1 {=}{{=\allowbreak}}1
  },
borderline west={2pt}{0pt}{blue!55!black}, left=5pt, right=5pt, top=5pt, bottom=5pt, before skip=0.8em, after skip=0.8em
}

\newtcolorbox{baselinepromptcontainerbox}[1]{enhanced jigsaw, breakable, colback=white, colframe=black!12, colbacktitle=black!85, coltitle=white, fonttitle=\bfseries\sffamily\small, title={#1}, boxrule=0.5pt, arc=1.4mm, outer arc=1.4mm, left=0.6mm, right=0.6mm, top=0.6mm, bottom=0.6mm, toptitle=1.4mm, bottomtitle=1.2mm, boxsep=0pt, before skip=0.55em, after skip=0.55em
}

\newenvironment{baselinepromptcontainer}[1]{%
\par\noindent\begin{baselinepromptcontainerbox}{#1}
}{%
  \end{baselinepromptcontainerbox}
}

\newlength{\baselinepromptindent}
\setlength{\baselinepromptindent}{0.38em}
\newcommand{\baselinepromptline}[1]{\noindent #1\par}
\newcommand{\baselinepromptindented}[2]{\noindent\hspace*{\dimexpr #1\baselinepromptindent\relax}#2\par}
\newcommand{\baselinepromptblank}{\vspace{0.12em}}

\colorlet{routercolor}{iconlearn}
\colorlet{specialistcolor}{iconfull}
\colorlet{validatorcolor}{deltacolor}
\colorlet{aggregatorcolor}{goldicon}
\colorlet{topologycolor}{black!50}

\newcommand{\baselineRoleIcon}[1]{%
  \ifstrequal{#1}{router}{\faRoute}{%
  \ifstrequal{#1}{specialist}{\faCogs}{%
  \ifstrequal{#1}{validator}{\faCheckCircle}{%
  \ifstrequal{#1}{aggregator}{\faFlagCheckered}{%
  \ifstrequal{#1}{topology}{\faSitemap}{\faUserCog}}}}}%
}

\newtcolorbox{baselinepartbox}[2]{enhanced jigsaw, breakable, colback=#1color!6, colframe=#1color!30, boxrule=0.35pt, arc=1.0mm, outer arc=1.0mm, borderline west={1.8pt}{0pt}{#1color!85!black}, before upper={%
    {\sffamily\bfseries\footnotesize\colorbox{#1color!18}{\textcolor{#1color!72!black}{\strut\hspace{0.4em}\baselineRoleIcon{#1}\hspace{0.45em}#2\hspace{0.55em}}}}\par
    \vspace{0.20em}%
    \ttfamily\scriptsize\linespread{1.05}\selectfont
    \setlength{\parindent}{0pt}%
    \setlength{\parskip}{0.10em}%
    \raggedright
  },
left=2.8mm, right=2.2mm, top=1.0mm, bottom=1.2mm, boxsep=0pt, pad at break=0.7mm, before skip=0.25em, after skip=0.28em
}

\noindent\textbf{Appendix Contents.}\par
\vspace{0.25em}
\begingroup
\small
\renewcommand{\arraystretch}{1.12}
\newcommand{\appentry}[3]{%
  \noindent\makebox[\columnwidth][l]{\hyperref[#2]{\textbf{#1\quad #3}}\hfill\pageref{#2}}\par}
\newcommand{\appsubentry}[3]{%
  \noindent\makebox[\columnwidth][l]{\hspace*{1.2em}\hyperref[#2]{#1.\quad #3}\hfill\pageref{#2}}\par}
\appentry{A}{app:extended_related}{Extended Related Work} \appsubentry{A.1}{app:rw_role_based}{Role-Based Multi-Agent Systems} \appsubentry{A.2}{app:rw_graph}{Graph-Structured Agent Collaboration} \appsubentry{A.3}{app:rw_pruning_credit}{Agent Pruning and Credit Assignment} \appsubentry{A.4}{app:rw_evolution}{Dynamic Role and Prompt Evolution}
\vspace{0.3em}

\appentry{B}{app:benchmark_details}{Benchmark and Evaluation Details} \appsubentry{B.1}{app:dataset_statistics}{Split and Subtask Statistics} \appsubentry{B.2}{app:metrics}{Evaluation Metrics}
\vspace{0.3em}

\appentry{C}{app:baselines}{Baseline Implementations} \appsubentry{C.1}{app:baseline_summary}{Overview} \appsubentry{C.2}{app:single_agent}{Single-Agent Baselines} \appsubentry{C.3}{app:fixed_workflow}{Workflow Baselines} \appsubentry{C.4}{app:expert_static_dag}{Static DAG MAS Baselines} \appsubentry{C.5}{app:static_seed}{Static Role Orchestration} \appsubentry{C.6}{app:random_evolution}{Random Role Evolution}
\vspace{0.3em}

\appentry{D}{app:sero_details}{\sero Implementation Details} \appsubentry{D.1}{app:controller_details}{Controller Details} \appsubentry{D.2}{app:hyperparameters}{Hyperparameters} \appsubentry{D.3}{app:seed_roles}{Seed Roles} \appsubentry{D.4}{app:training_strategy}{Training Strategy} \appsubentry{D.5}{app:credit_ranked_dag}{Credit-Ranked DAG Construction} \appsubentry{D.6}{app:training_loop}{Training Loop}
\vspace{0.3em}

\appentry{E}{app:additional_results}{Additional Results and Mechanism Analyses} \appsubentry{E.1}{app:per_seed}{Seed-Level Detailed Results} \appsubentry{E.2}{app:role_lifecycle_analysis}{Role-Pool Evolution} \appsubentry{E.3}{app:active_set_diversity}{Inference-Time Orchestration} \appsubentry{E.4}{app:credit_cost_analysis}{Credit and Topology Alignment} \appsubentry{E.5}{app:scaling}{Scaling Behavior} \par \endgroup

\vspace{0.8em}

\section{Extended Related Work}
\label{app:extended_related}

\subsection{Role-Based Multi-Agent Systems}
\label{app:rw_role_based}

Role-based LLM-agent systems decompose a task by assigning different prompts or personas to different model instances.  CAMEL studies role-playing societies of agents \citep{li2023camel}; MetaGPT turns software-development processes into coordinated agent roles \citep{hong2024metagpt}; AutoGen provides a framework for programmable multi-agent conversation \citep{Wu2023AutoGenEN}; and debate-style systems use multiple agents to expose and challenge intermediate reasoning \citep{du2024improving}.  These systems motivate \sero's use of roles, but they generally keep the role set fixed within a run or rely on developer-specified decompositions.  \sero instead treats the reusable role inventory as a state that can be edited under structural constraints.

\subsection{Graph-Structured Agent Collaboration}
\label{app:rw_graph}

Graph-structured agent methods focus on how agents communicate.  GPTSwarm optimizes collaboration graphs over predefined components \citep{zhuge2024gptswarm}; G-Designer and AMAS construct task-adaptive graphs \citep{zhang2025g,leong2025amas}; ARG-Designer jointly chooses agent roles and communication links for a query \citep{li2026assemble}; SC-MAS assigns edge-level collaboration strategies and model choices \citep{zhao2026sc}; GoAgent builds group-centric communication graphs \citep{chen2026goagent}; and TodyComm adapts communication across rounds \citep{fan2026todycomm}. These approaches are closely related because \sero also rebuilds a DAG for each active team.  The difference is that \sero makes graph construction dependent on typed role cards and persistent role-pool evolution, rather than only optimizing query-time communication structure or one-shot team composition.

\subsection{Agent Pruning and Credit Assignment}
\label{app:rw_pruning_credit}

Pruning and selection methods reduce the cost of multi-agent collaboration by removing unnecessary agents or links.  AgentPrune learns spatial-temporal masks for economical communication pipelines \citep{zhang2025cut}; AgentDropout removes redundant agents or links across collaboration rounds \citep{wang2025agentdropout}; AGP jointly prunes agent quantity and communication topology \citep{li2025adaptive}; and EIB-Learner analyzes how sparse and dense topologies propagate useful and erroneous information \citep{shen-etal-2025-understanding}.  Credit assignment provides complementary signals for estimating agent value, including difference rewards \citep{wolpert2001optimal}, Shapley-style values \citep{shapley201617}, and recent LLM-agent credit or contribution-guided optimization methods \citep{nagpal2025leveraging,NEURIPS2025_80164880,xia2026hivemind}.  \sero uses credit for ranking, routing, and controller state, but prevents credit noise from directly deleting protected terminal roles, validators, or the last role covering a required capability family.

\subsection{Dynamic Role and Prompt Evolution}
\label{app:rw_evolution}

Dynamic prompt and role systems aim to make multi-agent collaboration less dependent on hand-written decompositions.  Dynamic role assignment selects suitable debate participants for role slots \citep{zhang2026dynamic}; MetaGen rewrites query-conditioned role prompts and topologies at inference time \citep{wang2026metagen}; Multi-Agent Evolve studies self-play evolution of cooperative capabilities within a fixed proposer--solver--judge template \citep{chen2025multi}; and ARG-Designer can choose agents from an extensible pool as part of graph construction \citep{li2026assemble}.  \sero is closest to this family, but differs in how it treats adaptation: a candidate role-pool edit is not immediately trusted as a query-time assignment, prompt refinement, or graph decision.  It must be a schema-valid role-card update, satisfy diversity and benchmark constraints, preserve the structural contracts, and improve observed task score before it becomes part of the persistent pool.

\section{Benchmark and Evaluation Details}
\label{app:benchmark_details}

\subsection{Split and Subtask Statistics}
\label{app:dataset_statistics}

\textbf{Table~\ref{tab:split_subtask_stats}} summarizes the training and held-out evaluation partitions used in all experiments.  The training partition is reserved for trainable systems, and every method is evaluated on the same held-out test instances.  For \tablebench, the three task families aggregate the 17 non-visual question subtypes in the split.  For \olympiadbench, the raw benchmark pool is sampled from two text-only English open-ended competition subset: \texttt{OE\_TO\_maths\_en\_COMP} (674 items) and \texttt{OE\_TO\_physics\_en\_COMP} (236 items).  The effective split contains 897 available items rather than 910 because the loader applies an additional text-only guard and removes 13 entries that still reference visual material through words such as ``figure'', ``image'', or ``shown below'' (5 mathematics and 8 physics items).  No further \olympiadbench item is removed by the split quality filter.  The resulting available pool contains 669 mathematics and 228 physics items, from which we select 12/8 training examples and evaluate on the remaining 657/220 held-out examples.

\begin{table}[tbh]
\centering
\footnotesize
\setlength{\tabcolsep}{5pt}
\renewcommand{\arraystretch}{1.08}
\resizebox{\columnwidth}{!}{%
\begin{tabular}{llrr}
\toprule
Bench. & Subtask / family & Train & Held-out test \\
\midrule
NP & Trip planning & 10 & 300 \\
NP & Calendar scheduling & 10 & 300 \\
NP & Meeting planning & 10 & 300 \\
\cmidrule(lr){2-4}
NP & Total & 30 & 900 \\
\midrule
TB & Numerical reasoning & 19 & 378 \\
TB & Data analysis & 16 & 327 \\
TB & Fact checking & 5 & 91 \\
\cmidrule(lr){2-4}
TB & Total & 40 & 796 \\
\midrule
OB & Mathematics & 12 & 657 \\
OB & Physics & 8 & 220 \\
\cmidrule(lr){2-4}
OB & Total & 20 & 877 \\
\bottomrule
\end{tabular}
}
\caption{Training and held-out test composition by benchmark family and subtask.  NP, TB, and OB denote \naturalplan, \tablebench, and \olympiadbench.}
\label{tab:split_subtask_stats}
\end{table}

\subsection{Evaluation Metrics}
\label{app:metrics}

For benchmark $B$ with held-out set $\mathcal{D}_B=\{(x_i,y_i)\}_{i=1}^{N_B}$ and per-instance score $s_i=s_{x_i}(\hat y_i)\in[0,1]$, the mean task score is
\begin{equation}
  S_B=\frac{1}{N_B}\sum_{i=1}^{N_B} s_i,
  \label{eq:app_mean_score}
\end{equation}
where $\hat y_i$ is the system's parsed answer on task $x_i$ and $s_{x_i}(\cdot)$ is the benchmark-specific scorer.  For a subtask group $g$ with held-out instances $\mathcal{D}_{B,g}\subseteq\mathcal{D}_B$, the diagnostic score is
\begin{equation}
  S_{B,g}=\frac{1}{|\mathcal{D}_{B,g}|} \sum_{(x_i,y_i)\in \mathcal{D}_{B,g}} s_i .
  \label{eq:app_group_score}
\end{equation}

\paragraph{\naturalplan.}
\naturalplan records two per-instance metrics, a partial score $p_i\in[0,1]$ and an exact score $e_i\in\{0,1\}$.  With $N_{\mathrm{NP}}=|\mathcal{D}_{\mathrm{NP}}|$, we report
\begin{equation}
  P_{\mathrm{NP}}=\frac{1}{N_{\mathrm{NP}}}\sum_i p_i,\qquad E_{\mathrm{NP}}=\frac{1}{N_{\mathrm{NP}}}\sum_i e_i .
  \label{eq:app_np_metrics}
\end{equation}
For trip planning, the partial score is day-slot agreement.  Let $c_i^\star(t)$ be the gold city assigned to day $t$, $T_i$ the total number of trip days, and $\mathcal{Z}_i$ the set of non-empty parsed itinerary candidates extracted from the response.  Each candidate $z\in\mathcal{Z}_i$ induces a predicted day-city function $\hat c_{i,z}(t)$, and the trip partial score is
\begin{equation}
  p_i^{\mathrm{trip}}= \max_{z\in\mathcal{Z}_i} \frac{1}{T_i}\sum_{t=1}^{T_i} \mathbf{1}\!\left[\hat c_{i,z}(t)=c_i^\star(t)\right].
  \label{eq:app_trip_partial}
\end{equation}
Trip exact accuracy follows the NaturalPlan flight-sequence parser rather than raw-string equality.  Let $G_i=((c_{i1}^{\star},d_{i1}^{\star}),\ldots,(c_{iK_i}^{\star},d_{iK_i}^{\star}))$ be the gold city-duration sequence from the benchmark fields.  From a response, the parser extracts ordered flight statements, takes the first flight origin followed by each flight destination as the predicted city sequence $(\tilde c_{i1},\ldots,\tilde c_{i\tilde K_i})$, and infers inclusive stay lengths from the parsed flight days.  If $\tilde\tau_{i0}=1$, $\tilde\tau_{i\tilde K_i}$ is the end day of the final parsed visit range, and $\tilde\tau_{i1},\ldots,\tilde\tau_{i,\tilde K_i-1}$ are the extracted flight days, then
\begin{equation}
  \tilde d_{ik}=\tilde\tau_{ik}-\tilde\tau_{i,k-1}+1,
  \qquad k=1,\ldots,\tilde K_i.
  \label{eq:app_trip_exact_duration}
\end{equation}
The trip exact score is therefore
\begin{equation}
  e_i^{\mathrm{trip}}
  =\mathbf{1}\!\left[
    \tilde K_i\ge K_i\;\land\;
    \bigwedge_{k=1}^{K_i}
    \bigl((\tilde c_{ik},\tilde d_{ik})=(c_{ik}^{\star},d_{ik}^{\star})\bigr)
  \right],
  \label{eq:app_trip_exact}
\end{equation}
so any first mismatch in the ordered city-duration prefix gives zero exact credit.

For meeting planning, let $M_i^\star$ be the number of valid meetings in the gold route, $M_i^{\mathrm{partial}}$ the number of independently valid meetings counted by the permissive evaluator, and $M_i^{\mathrm{strict}}$ the strict valid-meeting count under break-on-first-error execution.  The scores are
\begin{equation}
\begin{aligned}
  p_i^{\mathrm{meet}} &=\min\!\left(1,\frac{M_i^{\mathrm{partial}}}{M_i^\star}\right),\\
  e_i^{\mathrm{meet}} &=\mathbf{1}\!\left[M_i^{\mathrm{strict}}=M_i^\star\right].
\end{aligned}
\label{eq:app_meeting_scores}
\end{equation}
The permissive evaluator skips invalid steps but advances time using the response's stated durations so later meetings can still receive credit.  The strict evaluator validates travel, waits, meeting windows, and stated times, and stops at the first error.  Calendar scheduling is the degenerate case where partial and exact coincide,
\begin{equation}
  p_i^{\mathrm{cal}}=e_i^{\mathrm{cal}}= \mathbf{1}\!\left[ \hat d_i=d_i^\star\land \hat s_i=s_i^\star\land \hat t_i=t_i^\star \right],
  \label{eq:app_calendar}
\end{equation}
with $d_i^\star, s_i^\star, t_i^\star$ the gold day, start time, and end time, and $\hat d_i, \hat s_i, \hat t_i$ their predicted counterparts after time normalization.

\paragraph{\tablebench.}
Each item receives lenient exact accuracy,
\begin{equation}
  s_i^{\mathrm{TB}}=\mathbf{1}\!\left[ M\bigl(\nu(\hat a_i),\nu(a_i^\star)\bigr) \right],
  \label{eq:app_tb_acc}
\end{equation}
where $\hat a_i$ and $a_i^\star$ are the predicted and gold answers, $\nu$ denotes answer normalization, and $M$ is exact string match except for supported numeric cases.  For tolerance-enabled statistical subtypes,
\begin{equation}
  M(\hat v,v)= \mathbf{1}\!\left[ \frac{|\hat v-v|}{\max(|v|,\epsilon)} \le 0.1 \right],
  \label{eq:app_tb_tolerance}
\end{equation}
with $\hat v$ and $v$ the predicted and gold numeric values and $\epsilon>0$ a small constant guarding against division by zero.

\paragraph{\olympiadbench.}
Let $J(\hat a,a^\star)$ be the OlympiadBench judge, with string matching as a fallback.  The benchmark contributes one scalar score per item.  For single-answer items, this score is binary,
\begin{equation}
  s_i^{\mathrm{OB}}=\mathbf{1}\!\left[J(\hat a_i,a_i^\star)\right],
  \label{eq:app_ob_single}
\end{equation}
where $\hat a_i$ and $a_i^\star$ are the predicted and gold answers.  For multi-answer items, the scorer first asks the judge to match the full predicted answer against the full gold answer; if this full-set check fails, it splits both sides into answer components and greedily matches each gold component to at most one predicted component.  With gold components $A_i^\star=\{a_{i1}^\star,\ldots,a_{iK}^\star\}$, predicted components $\hat A_i=\{\hat a_{i1},\ldots,\hat a_{iL}\}$, and $\mathcal{M}_{i}$ denoting the resulting one-to-one set of matched index pairs, the reported item score is
\begin{equation}
  s_i^{\mathrm{OB}} = \frac{1}{K}\sum_{(k,j)\in\mathcal{M}_{i}} \mathbf{1}\!\left[J(\hat a_{ij},a_{ik}^\star)\right].
  \label{eq:app_ob_multi}
\end{equation}

\section{Baseline Implementations}
\label{app:baselines}

This appendix defines the comparison systems used in the experiments.  All baselines are run on the same loaded tasks, benchmark adapters, canonical answer extractors, and scoring functions as \sero.  Let \(x_i\) denote a task, \(F_b\) the benchmark-specific answer extractor, and \(S_b\) the scorer.  For a multi-agent baseline with role cards \(R_b\), directed message edges \(E_b\), and terminal role \(t_b\), execution is a topological pass over a fixed graph, with \(U_r=\{m_u:(u,r)\in E_b\}\), \(m_r=f_\theta(\pi_r,x_i,U_r)\), and \(\hat{a}_i=F_b(m_{t_b})\). where \(\pi_r\) is the system prompt stored in role card \(r\).  The baselines differ only in how the role set and graph are chosen, and in whether the role pool is allowed to change.  Detailed prompt bundles for the multi-agent baselines appear in \Cref{app:fixed_workflow,app:expert_static_dag,app:seed_roles}.

\subsection{Overview}
\label{app:baseline_summary}

This subsection summarizes what each baseline fixes or learns and the contribution it isolates. Detailed implementations are in \Cref{app:single_agent,app:fixed_workflow,app:expert_static_dag,app:static_seed,app:random_evolution}.

\begin{itemize}[leftmargin=*,itemsep=2pt,topsep=2pt]
\item \textbf{CoT.} A single deterministic response from the base model at temperature \(0\) using the same benchmark-specific single-agent prompt as \sero, which measures strong single-agent prompting without multi-agent structure.
\item \textbf{SC@3.} Three temperature-\(0.7\) samples followed by majority vote over canonical extracted answers, which tests whether simple sampling can substitute for role orchestration.
\item \textbf{Workflow.} A benchmark-specific hand-written linear decomposition with no credit, retrieval, DAG construction, or evolution, which controls for the contribution of any sequentially-decomposed pipeline regardless of structure.
\item \textbf{Static DAG MAS.} A benchmark-specific hand-designed non-linear role graph with a fixed topology and fixed terminal aggregator, which tests whether expert-designed topology alone can explain the performance attributed to learned coordination.
\item \textbf{Static Role Orchestration.} \sero's seed pool combined with the same retrieval, credit-ranked DAG execution, aggregation, and validator repair pipeline, but with role-pool evolution disabled, which isolates the value of learned role-pool evolution beyond the rest of the inference pipeline.
\item \textbf{Random Role Evolution.} The same edit space and role-card editor as \sero, but edit operations are sampled uniformly at random instead of from the learned controller and are committed without a score gate, which tests whether the controller and commitment rule add value beyond unconstrained exploration.
\item \textbf{\sero (Ours).} The learned guarded controller, constrained role editor, score-gated commitment, and frozen test-time pool, evaluated under the same inference operator as the static-pool baselines.
\end{itemize}

\subsection{Single-Agent Baselines}
\label{app:single_agent}

CoT \citep{wei2022chain} is the deterministic one-call baseline.  For benchmark \(b\), let \(p_b^{\mathrm{SA}}\) denote the benchmark-specific single-agent system prompt used throughout the evaluation protocol.  The evaluator queries the base model once with temperature \(T=0\): \[ y_i = f_\theta(p_b^{\mathrm{SA}}, x_i; T=0), \qquad \hat{a}_i=F_b(y_i). \] SC@3 \citep{wang2022self} keeps the same prompt but samples three independent responses at temperature \(0.7\), extracts a canonical answer from each response, and returns the majority answer:
\[
\begin{aligned}
  y_i^{(j)} &\sim f_\theta(p_b^{\mathrm{SA}},x_i;T=0.7),\qquad j=1,2,3,\\
  z_i^{(j)} &= F_b(y_i^{(j)}),\\
  \hat{a}_i
  &= \operatorname*{arg\,max}_{\substack{z\in\{z_i^{(1)},z_i^{(2)},z_i^{(3)}\}\\ z\ne\emptyset}}
     \sum_{j=1}^{3}\mathbf{1}\!\left[z_i^{(j)}=z\right].
\end{aligned}
\]
Ties in this vote are resolved by the earliest sampled response.  If every response fails canonical extraction, SC@3 scores the first sampled response.  This makes the comparison depend on the same answer normalization as \sero, rather than on free-form text similarity.


\subsection{Workflow Baselines}
\label{app:fixed_workflow}

Workflow is a hand-written sequential multi-agent baseline.  For each supported benchmark, the baseline uses a fixed ordered list of role cards and a single terminal card.  The graph is always a chain, \(r_1 \rightarrow r_2 \rightarrow \cdots \rightarrow r_K=t_b\), so execution reduces to \(m_{r_1}=f_\theta(\pi_{r_1},x_i)\) and \(m_{r_k}=f_\theta(\pi_{r_k},x_i,m_{r_{k-1}})\) for \(k=2,\ldots,K\). This baseline only passes messages and calls agents.  It does not use the adaptive retrieval module, the learned controller, credit computation, role selection, or role-pool evolution.

\begin{figure*}[tbh]
\centering
\includegraphics[width=\textwidth]{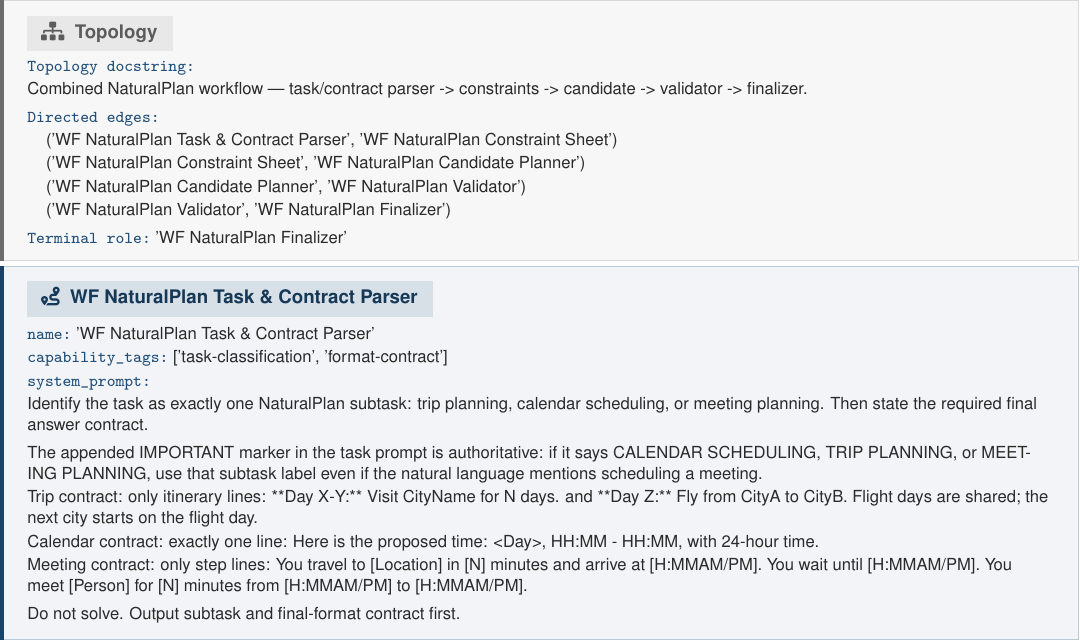}
\caption{Workflow baseline for \naturalplan~(part 1/3): topology and task-and-contract parser.}
\label{fig:wf_np_1}
\end{figure*}

\begin{figure*}[tbh]
\centering
\includegraphics[width=\textwidth]{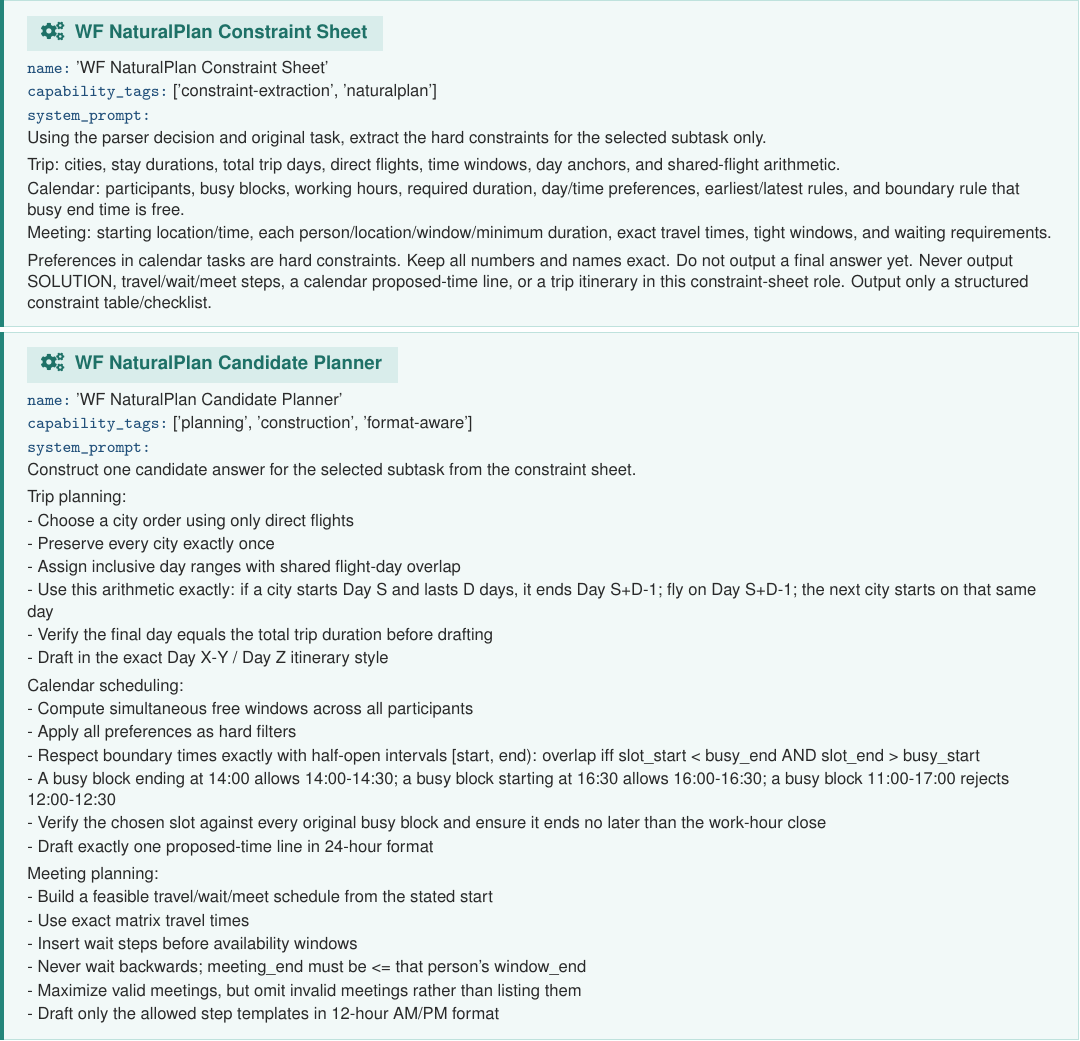}
\caption{Workflow baseline for \naturalplan~(part 2/3): constraint sheet and candidate planner.}
\label{fig:wf_np_2}
\end{figure*}

\begin{figure*}[tbh]
\centering
\includegraphics[width=\textwidth]{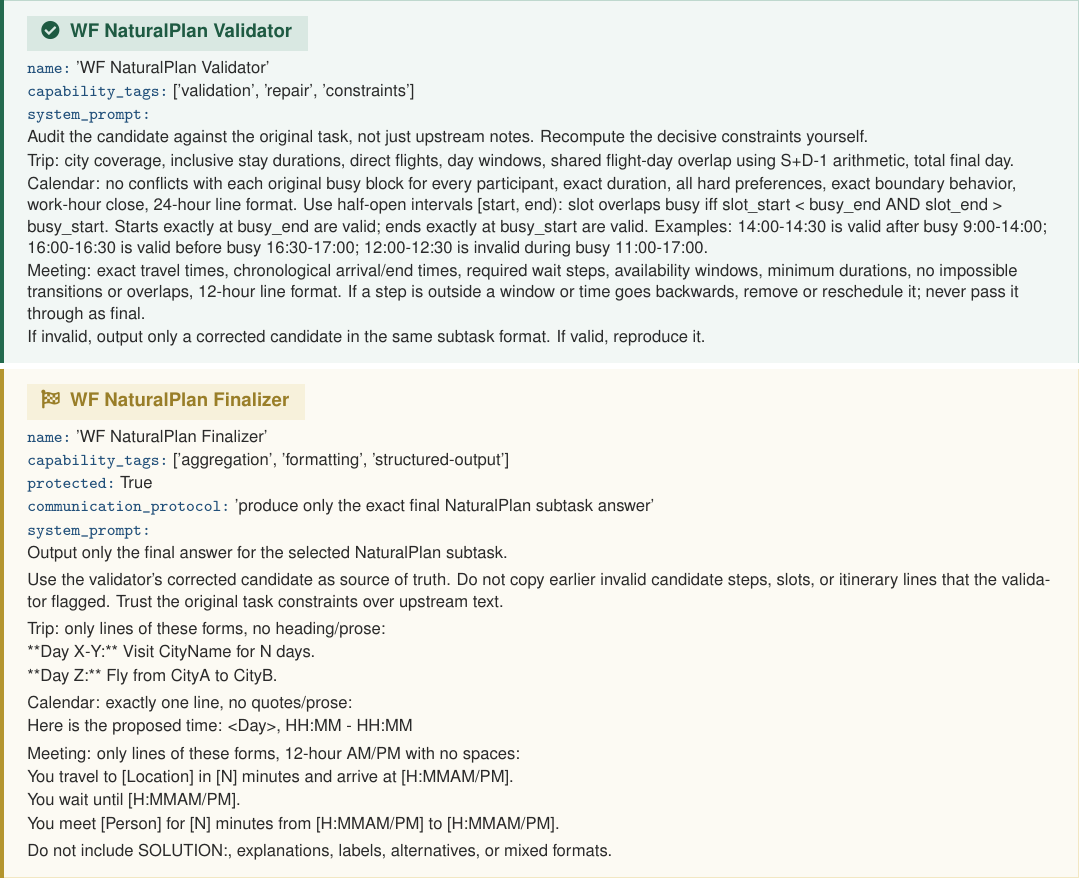}
\caption{Workflow baseline for \naturalplan~(part 3/3): validator and finalizer.}
\label{fig:wf_np_3}
\end{figure*}

\textbf{Figs.~\ref{fig:wf_np_1}--\ref{fig:wf_np_3}} give the full Workflow specification for \naturalplan: the topology, role-card fields, and exact system-prompt content of every role in the chain.  \olympiadbench uses an analogous \emph{problem decomposer $\to$ solution strategist $\to$ verification expert $\to$ answer synthesizer} chain with mathematics and physics specialists in place of trip / calendar / meeting roles, and \tablebench follows an analogous \emph{table parser $\to$ evidence selector $\to$ table solver $\to$ answer auditor $\to$ final formatter} chain over tabular questions.

\subsection{Static DAG MAS Baselines}
\label{app:expert_static_dag}

Static DAG MAS is a hand-designed non-linear multi-agent baseline.  Each benchmark receives a dedicated role set, a fixed edge set, and a fixed terminal role.  Unlike Workflow, a role may receive several upstream messages, and independent branches are executed in parallel levels.  If \(L_1,\ldots,L_H\) are the resulting topological levels, then for each level \[
\begin{aligned}
  U_r &= \{m_u:(u,r)\in E_b\},\\
  m_r &= f_\theta(\pi_r,x_i,U_r),\qquad r\in L_h.
\end{aligned}
\] The terminal role then receives its predecessors and produces the final answer. This baseline has no learned controller, no credit ranking, no retrieval over a seed pool, and no role-pool edits.

\begin{figure*}[tbh]
\centering
\includegraphics[width=\textwidth]{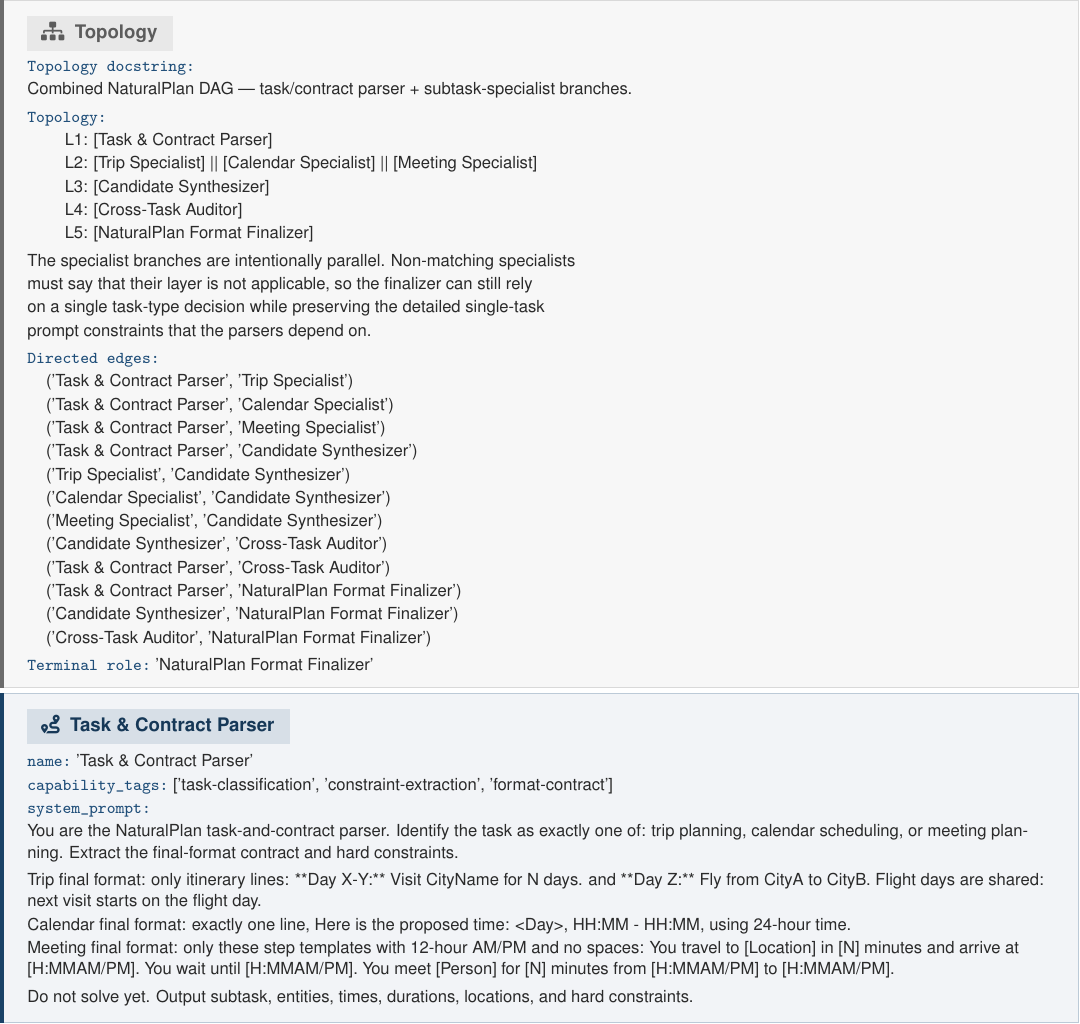}
\caption{Static DAG MAS baseline for \naturalplan~(part 1/4): topology and task-and-contract parser.}
\label{fig:dag_np_1}
\end{figure*}

\begin{figure*}[tbh]
\centering
\includegraphics[width=\textwidth]{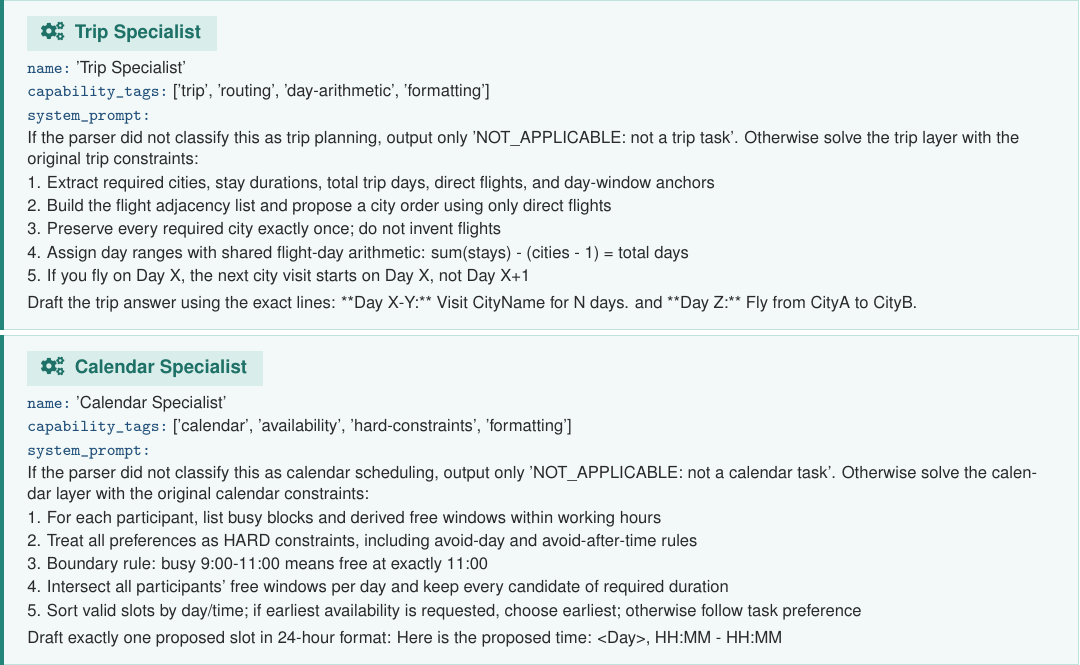}
\caption{Static DAG MAS baseline for \naturalplan~(part 2/4): trip and calendar specialists.}
\label{fig:dag_np_2}
\end{figure*}

\begin{figure*}[tbh]
\centering
\includegraphics[width=\textwidth]{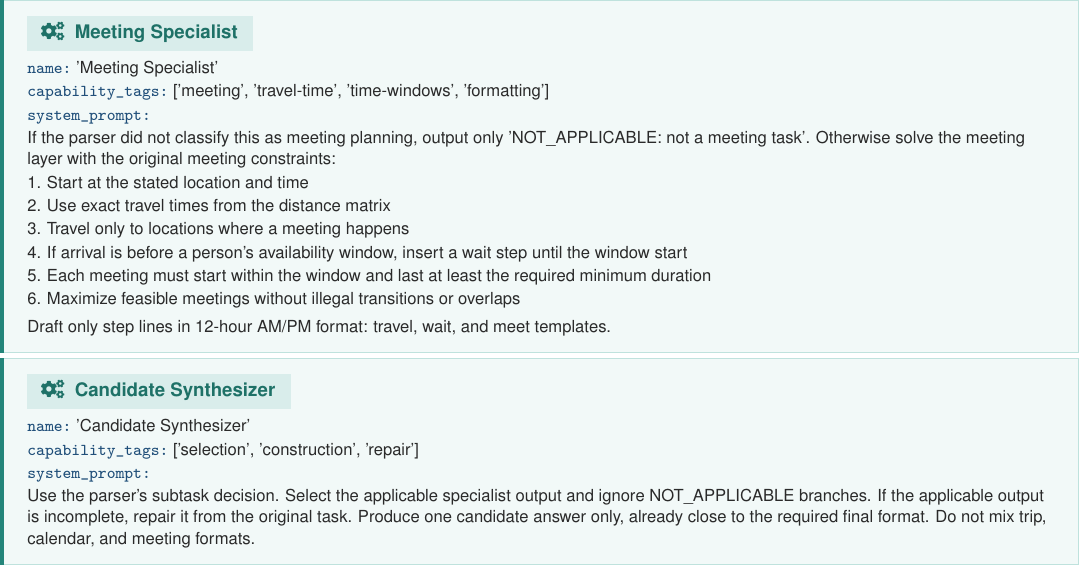}
\caption{Static DAG MAS baseline for \naturalplan~(part 3/4): meeting specialist and candidate synthesizer.}
\label{fig:dag_np_3}
\end{figure*}

\begin{figure*}[tbh]
\centering
\includegraphics[width=\textwidth]{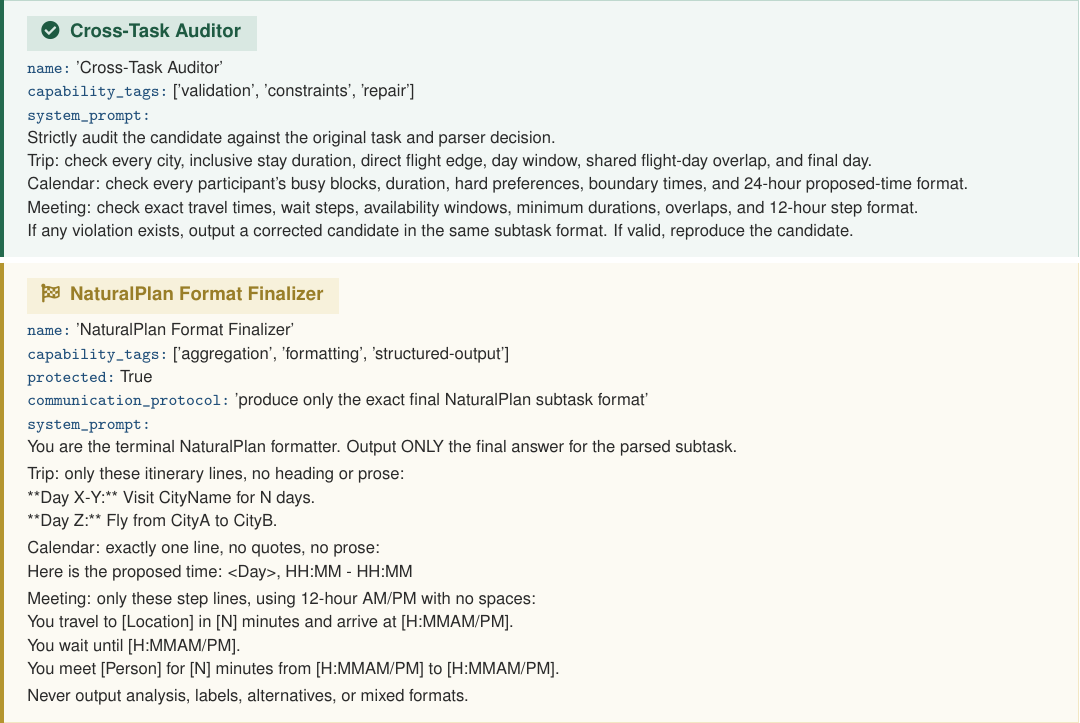}
\caption{Static DAG MAS baseline for \naturalplan~(part 4/4): cross-task auditor and protected format finalizer.}
\label{fig:dag_np_4}
\end{figure*}

\textbf{Figs.~\ref{fig:dag_np_1}--\ref{fig:dag_np_4}} give the full Static DAG specification for \naturalplan: the parallel trip / calendar / meeting specialist branches, the cross-task auditor, and the protected format finalizer.  \olympiadbench uses a \emph{concept identifier $\to$ algebraic / intuitive solver branches $\to$ cross-checker $\to$ answer extractor} DAG, and \tablebench uses a \emph{question schema mapper $\to$ evidence retriever $\to$ numerical / lookup specialist branches $\to$ arithmetic auditor $\to$ answer-auditor synthesizer $\to$ final formatter} DAG with the parser-facing format enforced by the protected terminal.

\subsection{Static Role Orchestration}
\label{app:static_seed}

Static Role Orchestration is the frozen seed-pool variant of \sero, not the expert static graph above: it uses \sero's benchmark seed role pool and the same retrieval-and-DAG inference engine as \sero.  The only disabled component is role-pool evolution. Formally, if \(\mathcal{P}_0\) is the seed pool, then \[
\begin{aligned}
  \mathcal{P}_{i+1} &= \mathcal{P}_i = \mathcal{P}_0,\\
  A_i &= \operatorname{Retrieve}(x_i,\mathcal{P}_0),\\
  G_i &= \operatorname{DAG}(A_i).
\end{aligned}
\] The static seed roles are still selected, ordered, aggregated, and optionally repaired by the same inference machinery as \sero, using current-task bootstrap signals to order the DAG but without updating fast-credit state during evaluation.  No \textsc{Add}, \textsc{Remove}, or \textsc{Noop} action can alter the pool.

\subsection{Random Role Evolution}
\label{app:random_evolution}

Random Role Evolution isolates whether performance gains can be explained by role-pool perturbation alone.  It starts from the same seed pool as \sero and uses the same role-card editor, schema validation, maximum pool size $P_{\max}$, protected-role removal constraint, and frozen evaluation operator.  Its only difference from \sero's learned evolution stage is that edit choices are sampled uniformly and are never accepted or rejected by a before--after score comparison.  For training task $x_i$ and current pool $\mathcal{P}_i$,
\[
\begin{aligned}
  \mathcal{O} &= \{\textsc{Add},\textsc{Remove},\textsc{Noop}\},\\
  \omega_i &\sim \operatorname{Unif}(\mathcal{O}),\\
  \rho_i &\sim \operatorname{Unif}(\mathcal{P}_i)
  \quad \text{if } \omega_i\ne\textsc{Noop},\\
  \mathcal{P}_{i+1} &= U(\mathcal{P}_i,\omega_i,\rho_i;x_i).
\end{aligned}
\]
Here $\rho_i$ denotes the sampled target role and $U$ denotes the random role-pool update: \textsc{Remove} deletes $\rho_i$ unless doing so would remove a protected role or empty the pool, \textsc{Add} appends a schema-valid role proposed from the current task context and the role with highest historical EMA credit when $|\mathcal{P}_i|<P_{\max}$, and \textsc{Noop} leaves the pool unchanged.  Scores observed on the training tasks are recorded only for analysis, not for edit commitment.  After the random evolution pass, the final pool is frozen and evaluated with the same retrieval, bootstrap credit-ranked DAG construction, aggregation, and validator-repair procedure used for \sero.

\section{\sero Training Details}
\label{app:sero_details}

This section documents the controller specification, the hyperparameter profiles, the seed role cards, the training strategy, the credit-ranked DAG construction routine, and the full training step used in the experiments.

\subsection{Controller Details}
\label{app:controller_details}

The controller is the only trainable neural component in \sero.  Intuitively, it reads three pieces of state: how the current task attempt interacted with the benchmark, which roles were active in that attempt, and how healthy the current role pool looks according to the five credit statistics in \Cref{eq:credit_state_summary}.  Under the 512-dimensional encoder used in these experiments, these signals form a 1029-dimensional state vector.  A shared two-layer MLP with hidden size 256 maps this vector to a latent state.  One head chooses the edit type from $\{\textsc{Add},\textsc{Remove},\textsc{Noop}\}$, while a second conditional head scores admissible target roles using the latent state, a learned 64-dimensional operation embedding, a projected role embedding, and two local credit features: the role's historical EMA credit and recent leave-one-out credit.  We train this policy with REINFORCE using batch-normalized rewards, an EMA baseline, and entropy regularization, while keeping both the task backbone and the text encoder frozen.

\subsection{Hyperparameters}
\label{app:hyperparameters}

The experiments use a small family of closely related profiles rather than a single universal setting.  \textbf{Table~\ref{tab:hyperparams}} lists the reference profile for Qwen-3-8b together with the alternative profile used for GPT-4o-mini and Gemini-2.5-flash-lite.  Main comparisons average runs over seeds 42, 43, and 44.  Hyperparameter sensitivity, ablation, and controller-scaling experiments use seed 44, while base-model scaling uses seed 43.  Unless a dedicated sensitivity analysis explicitly varies a given factor, both the hyperparameter-sensitivity experiments and the ablation experiments inherit the same settings as the corresponding main experiment.  Scaling experiments follow the corresponding main-experiment model--benchmark profile, except that the Qwen3 model-size sweep uses the GPT-4o-mini / Gemini profile in \textbf{Table~\ref{tab:hyperparams}} for all Qwen3 sizes to isolate backbone scale under a fixed training configuration.
\begin{table}[tbh]
\centering
\small
\resizebox{\columnwidth}{!}{%
\begin{tabular}{lcc}
\toprule
Parameter & Qwen-3-8b & GPT-4o-mini / Gemini \\
\midrule
Warmup / main epochs & 2 / 8 & 1 / 9 \\
Batch size & 8 & 4 \\
Collaboration rounds & 1 & 1 \\
\texttt{n\_max} & 4 & 4 \\
Specialist / validator slots & 3 / 1 & 5 / 1 \\
Maximum / minimum pool size & 12 / 4 & 10 / 3 \\
LOO refresh interval & 40 ep. & 20 ep. \\
LOO minimum pool size & 4 & 4 \\
New-role initial updates & 3 & 3 \\
Entropy coefficient & 0.05 & 0.08 \\
Learning rate & 0.001 & 0.001 \\
Historical-credit EMA decay $\mu$ & 0.1 & 0.05$^{\dagger}$ \\
Fast-credit $\alpha$ & 0.5 & 0.5 \\
Exploration $\gamma$ & 0.1 & 0.15 \\
Collapse threshold & 0.85 / 24 ep. & 0.85 / 24 ep. \\
\bottomrule
\end{tabular}
}
\caption{\sero hyperparameter profiles for the three backbones.  $^{\dagger}$For Gemini-2.5-flash-lite on \naturalplan, $\mu=0.2$ instead of $0.05$.}
\label{tab:hyperparams}
\end{table}

\subsection{Seed Roles}
\label{app:seed_roles}

The main three-benchmark setting starts from a 10-role cross-task \naturalplan pool, a 7-role \tablebench pool, and a 7-role \olympiadbench pool.  The highlighted blocks below reproduce the current seed role cards, including role type, protection flag, capability family, communication protocol, and the exact system-prompt line structure.  Code-level \texttt{role\_type} values \texttt{'router'}, \texttt{'specialist'}, \texttt{'validator'}, and \texttt{'aggregator'} correspond respectively to the canonical role types $\mathsf{Setup}$, $\mathsf{Spec}$, $\mathsf{Val}$, and $\mathsf{Agg}$ from \Cref{sec:role_cards}.  The single-column prompt blocks below are presented in the order Combined \naturalplan, \tablebench, and \olympiadbench.

\begin{figure*}[tbh]
\centering
\includegraphics[width=\textwidth]{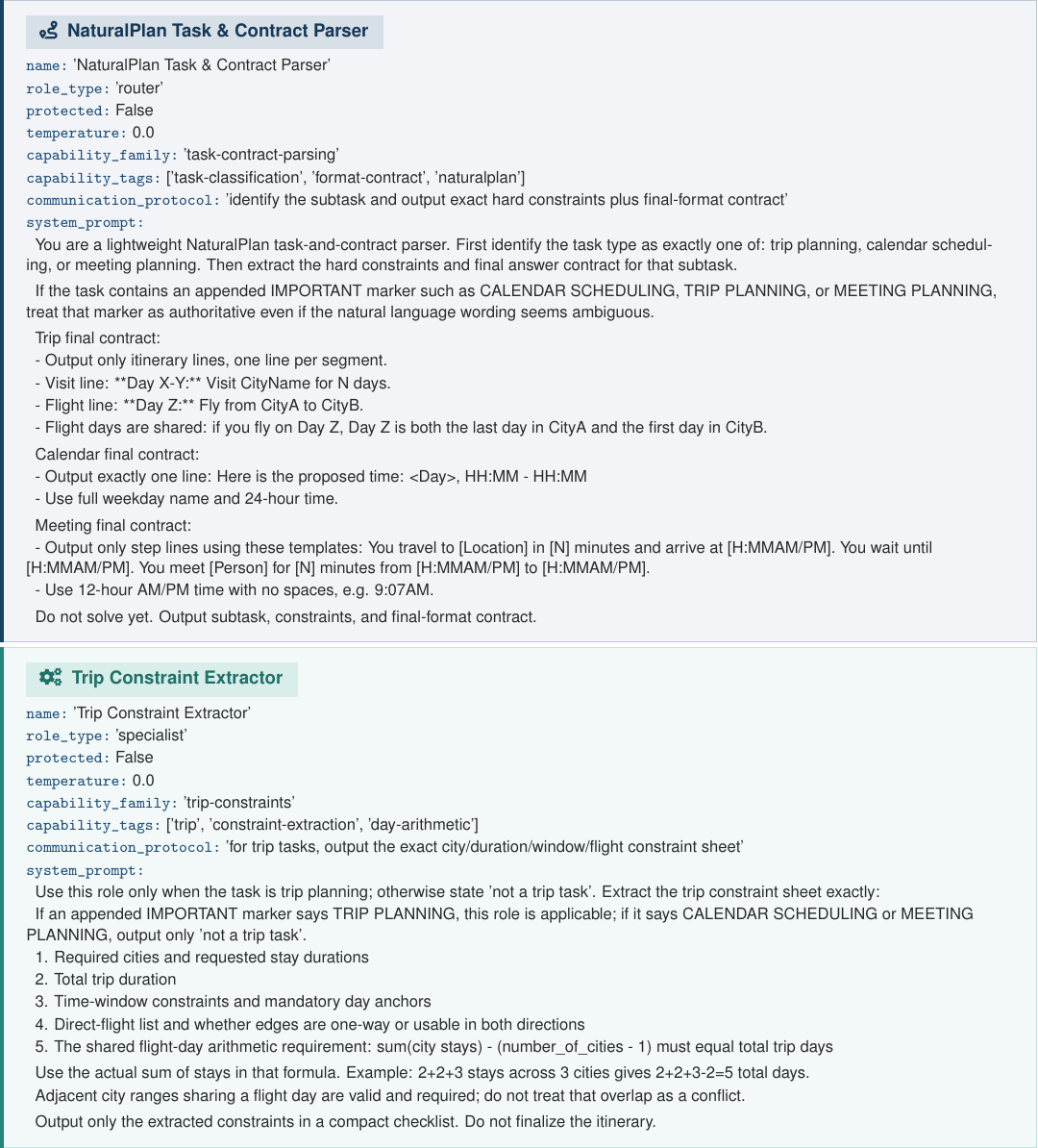}
\caption{\naturalplan seed pool~(part 1/6): task-and-contract parser and trip constraint extractor.}
\label{fig:seed_np_1}
\end{figure*}
\begin{figure*}[tbh]
\centering
\includegraphics[width=\textwidth]{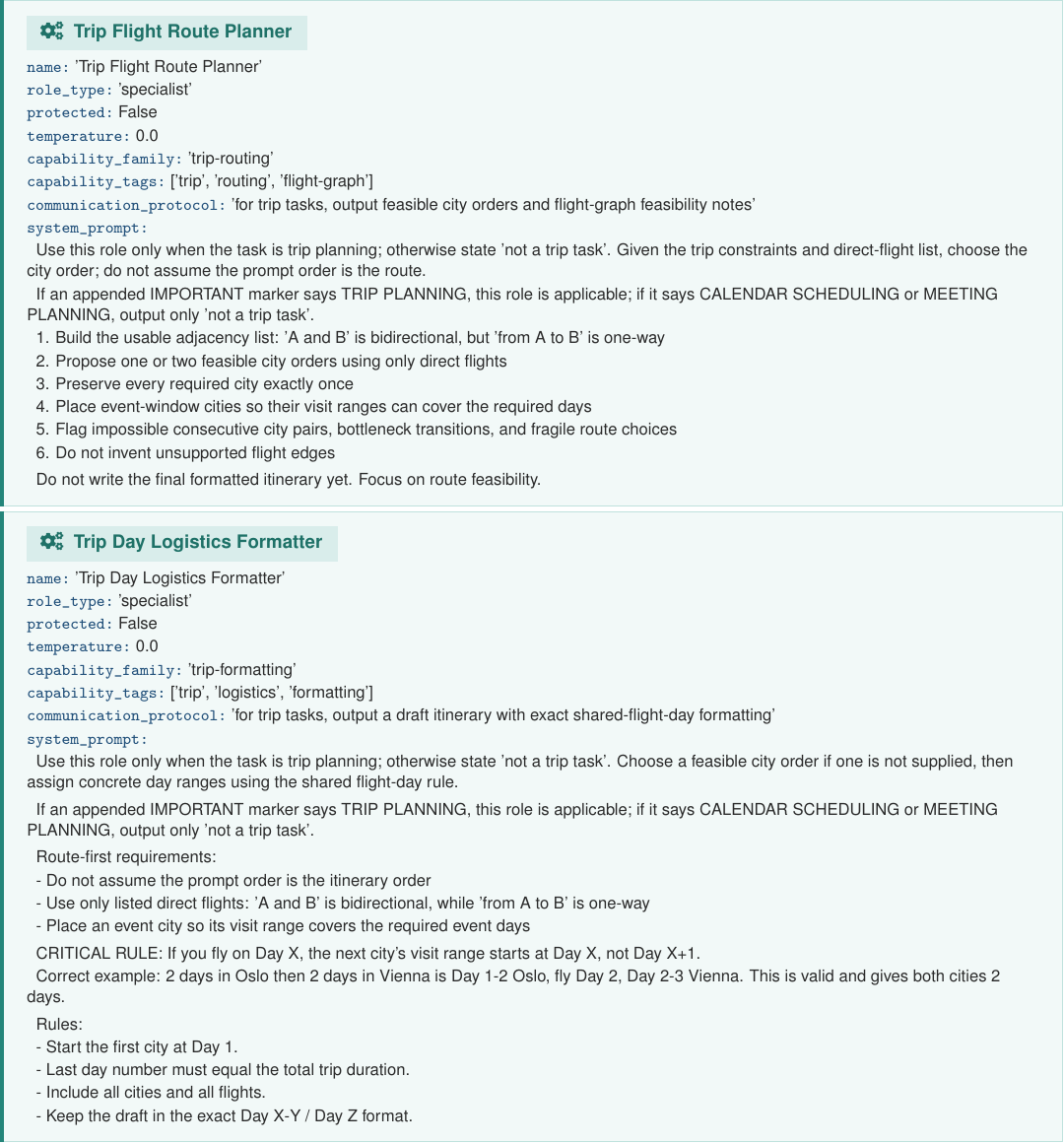}
\caption{\naturalplan seed pool~(part 2/6): trip flight route planner and trip day logistics formatter.}
\label{fig:seed_np_2}
\end{figure*}
\begin{figure*}[tbh]
\centering
\includegraphics[width=\textwidth]{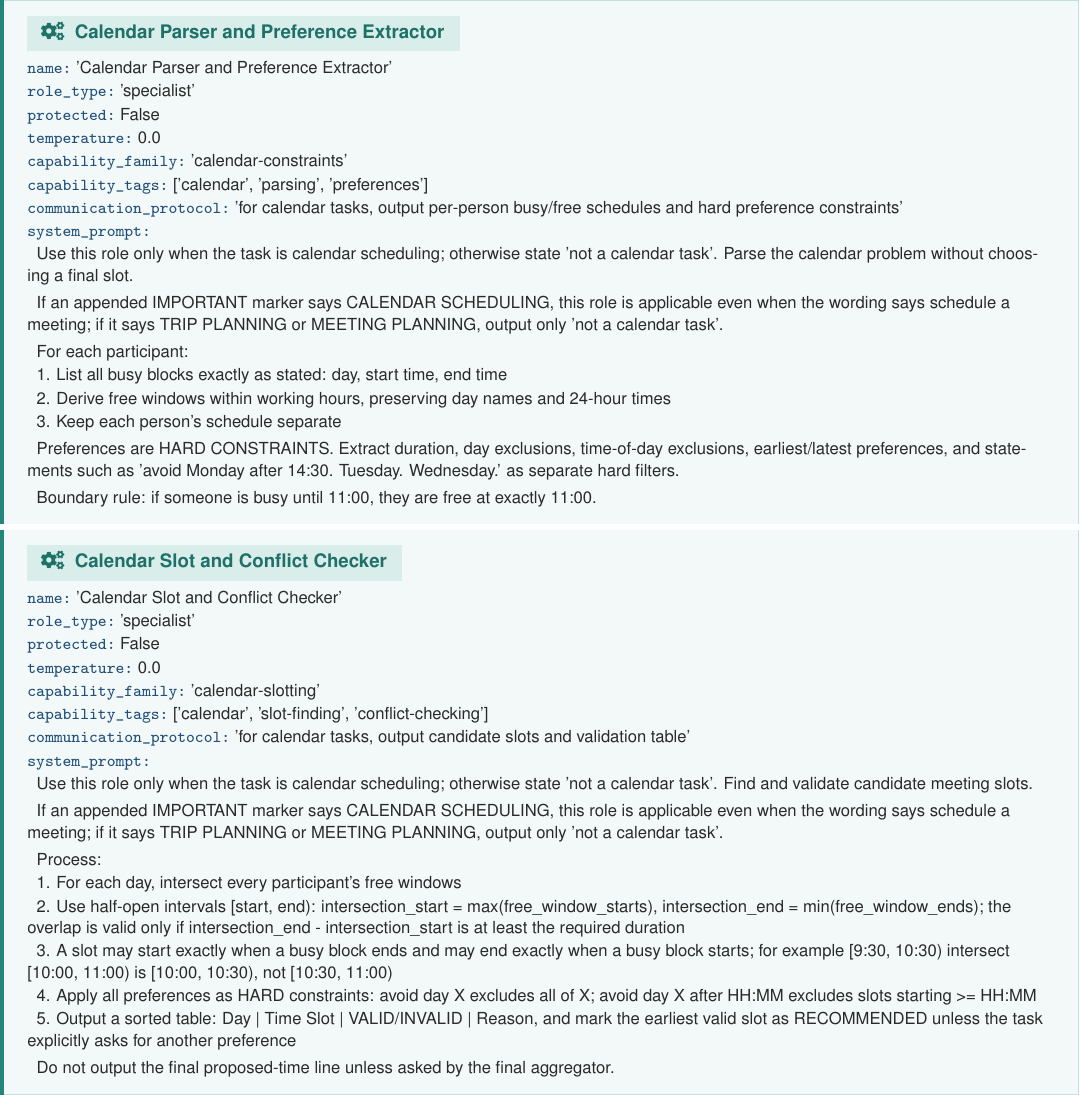}
\caption{\naturalplan seed pool~(part 3/6): calendar parser and calendar slot checker.}
\label{fig:seed_np_3}
\end{figure*}
\begin{figure*}[tbh]
\centering
\includegraphics[width=\textwidth]{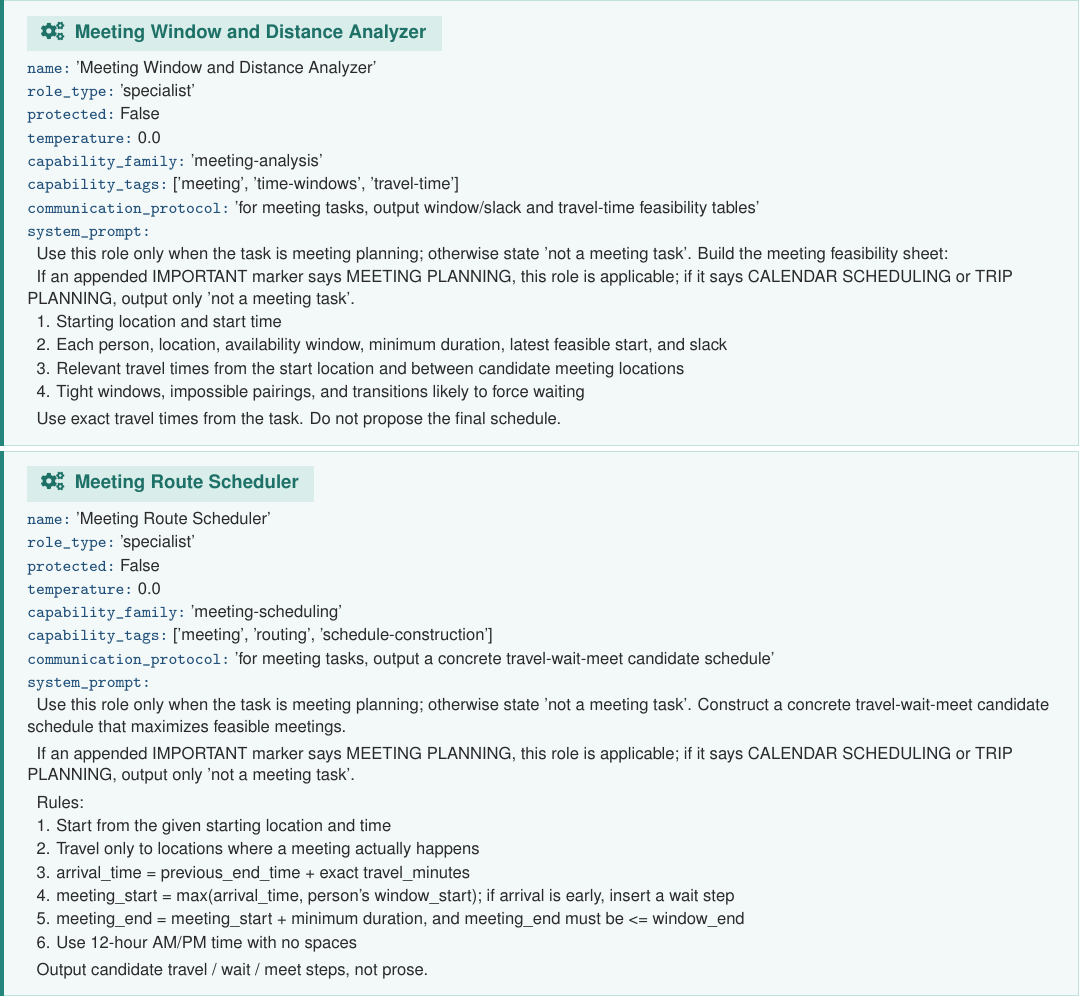}
\caption{\naturalplan seed pool~(part 4/6): meeting window analyzer and meeting route scheduler.}
\label{fig:seed_np_4}
\end{figure*}
\begin{figure*}[tbh]
\centering
\includegraphics[width=\textwidth]{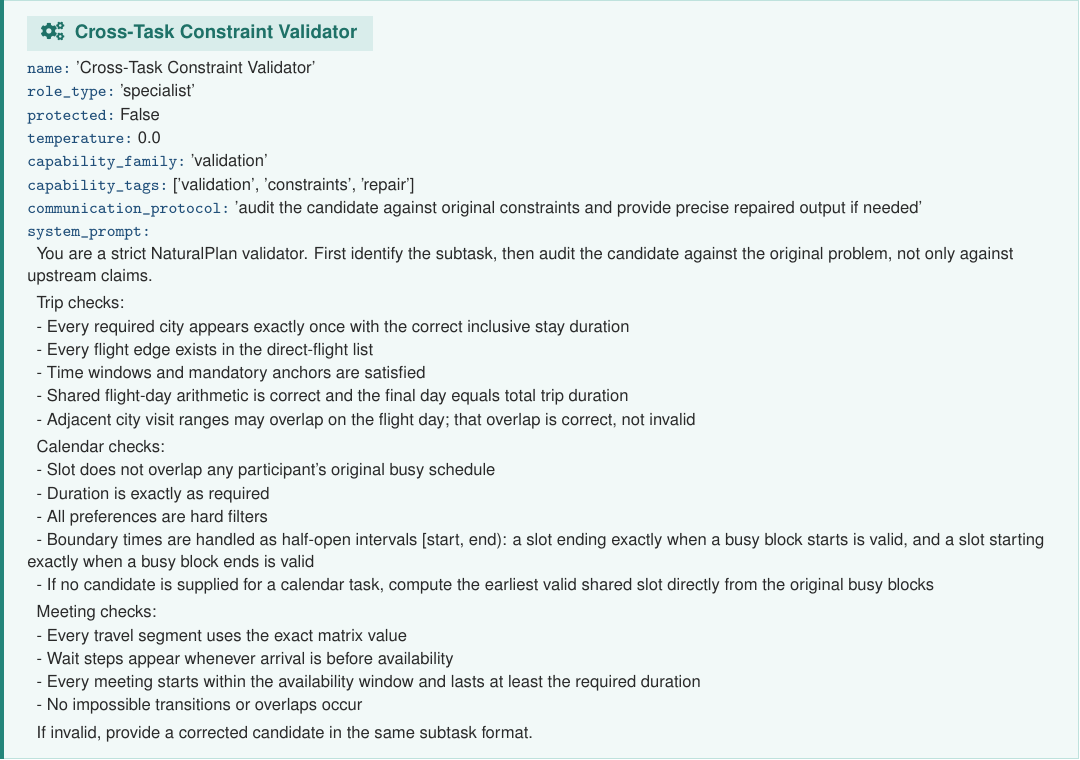}
\caption{\naturalplan seed pool~(part 5/6): cross-task constraint validator.}
\label{fig:seed_np_5}
\end{figure*}
\begin{figure*}[tbh]
\centering
\includegraphics[width=\textwidth]{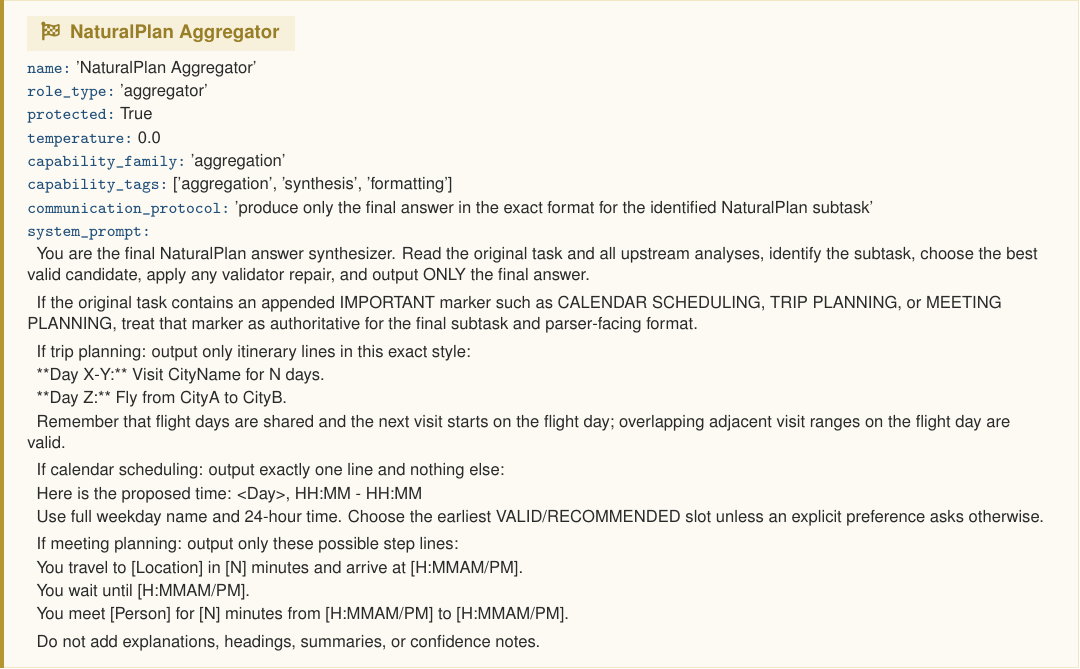}
\caption{\naturalplan seed pool~(part 6/6): protected \naturalplan aggregator.}
\label{fig:seed_np_6}
\end{figure*}
\begin{figure*}[tbh]
\centering
\includegraphics[width=\textwidth]{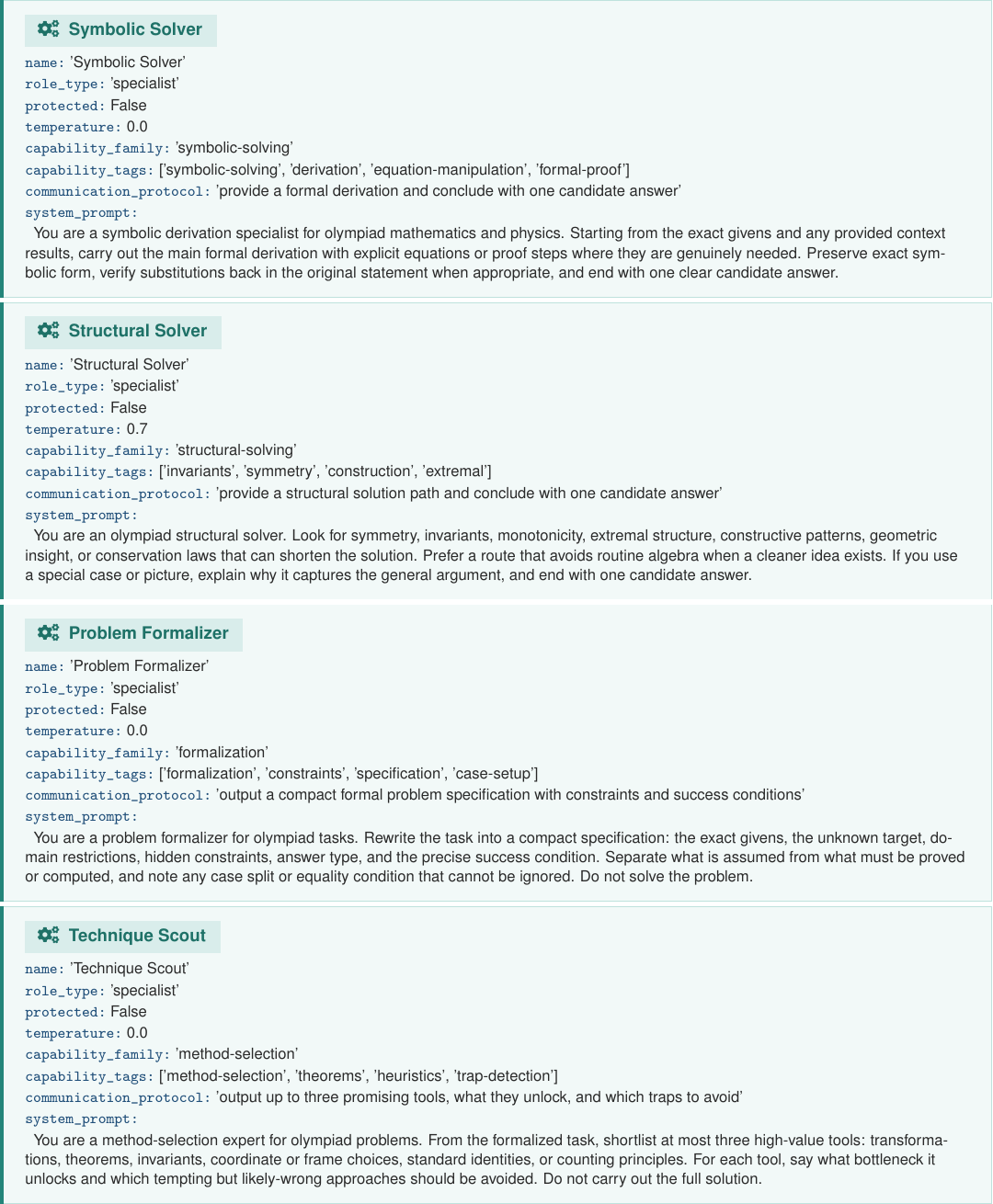}
\caption{\olympiadbench seed pool~(part 1/2): symbolic solver, structural solver, problem formalizer, and technique scout.}
\label{fig:seed_ob_1}
\end{figure*}
\begin{figure*}[tbh]
\centering
\includegraphics[width=\textwidth]{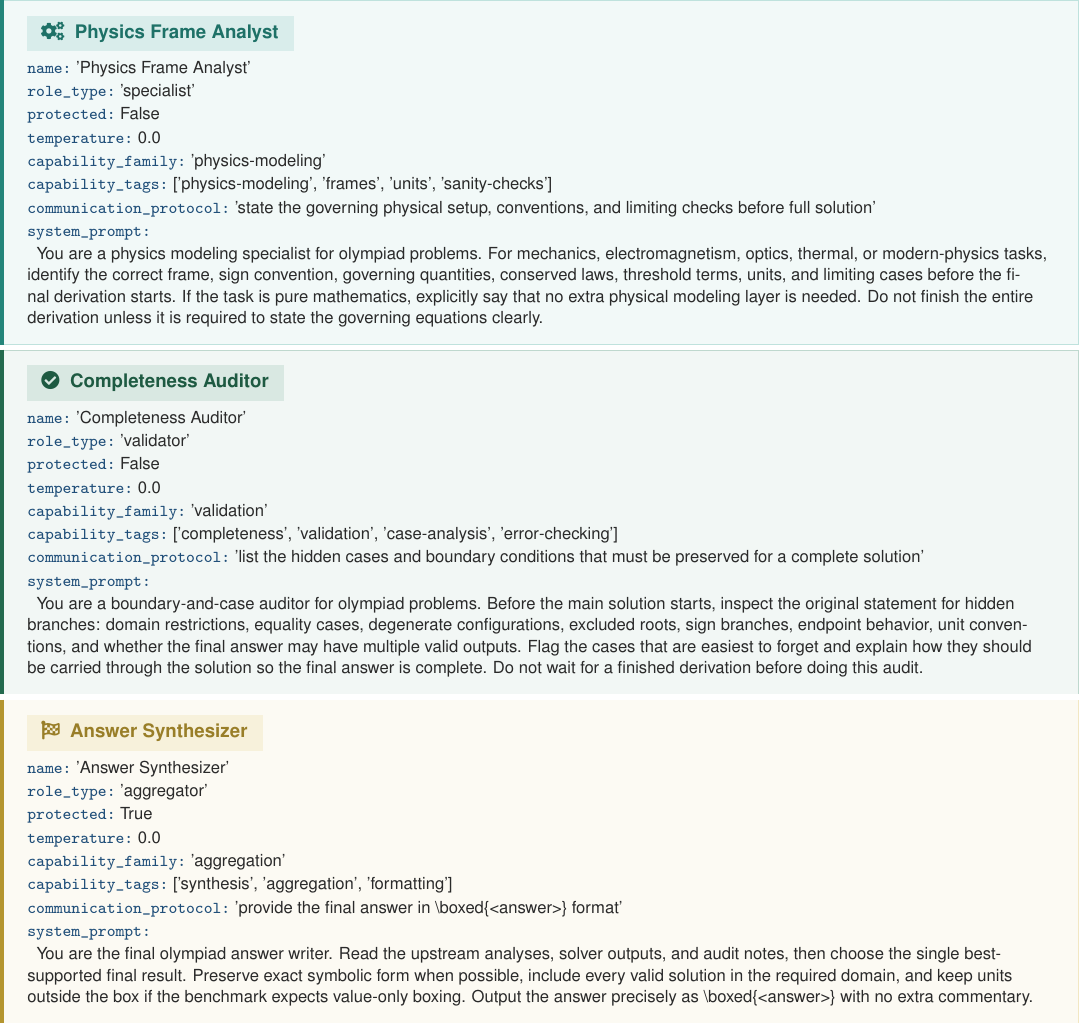}
\caption{\olympiadbench seed pool~(part 2/2): physics frame analyst, completeness auditor, and protected answer synthesizer.}
\label{fig:seed_ob_2}
\end{figure*}

\textbf{Figs.~\ref{fig:seed_np_1}--\ref{fig:seed_np_6}} show the ten role cards of the \naturalplan seed pool: a task-and-contract parser, eight domain specialists covering trip, calendar, and meeting subtasks plus a cross-task constraint validator, and a protected aggregator that enforces the parser-facing final format.  \textbf{Figs.~\ref{fig:seed_ob_1} and \ref{fig:seed_ob_2}} show the seven role cards of the \olympiadbench seed pool: symbolic and structural solvers, a problem formalizer, a technique scout, a physics frame analyst, a completeness auditor, and a protected answer synthesizer.  The \tablebench seed pool follows the same router-specialist-validator-aggregator structure with seven roles spanning schema mapping, evidence retrieval, numerical reasoning, data-analysis interpretation, fact checking, an answer verifier, and a protected aggregator.

\subsection{Training Strategy}
\label{app:training_strategy}
\label{app:training_protocol}
\label{app:guards}

Training is one-edit-per-task policy learning over a committed role pool. Given task $x_i$ and committed state $(\roles_i,C_i)$, the inference operator first produces a pre-edit answer $\hat y_i$, score $s_i^{-}=s_{x_i}(\hat y_i)$, and controller observation $o_i$.  The controller then samples an edit action $a_i=(\omega_i,z_i)$ with $\omega_i\in\mathcal{O}=\{\textsc{Add},\textsc{Remove},\textsc{Noop}\}$, where $z_i$ is a target role when the action requires one.  The candidate generator $\mathcal{E}$ proposes a candidate state $(\widetilde{\roles}_i,\widetilde C_i)$, and the same task is re-evaluated under this candidate state to produce a post-edit score $s_i^{+}$.  The episode reward is $R_i=0$ for \textsc{Noop} or an unchanged candidate, and $R_i=s_i^{+}-s_i^{-}$ otherwise.

The training schedule has a warmup phase followed by a main phase.  Warmup omits credit-state features from the controller input and disables removal, so the pool can only add roles or stay unchanged.  In warmup, non-NaturalPlan real edits are committed when they are non-harmful ($R_i\ge 0$); in the main phase, committed edits must be strictly improving ($R_i>0$).  NaturalPlan uses strict add acceptance throughout, so newly proposed roles are committed only when they improve the task score.  Rejected real edits restore the pre-edit credit state exactly, which prevents tentative fast-credit updates from leaking into later episodes.

Each training batch applies REINFORCE with batch-normalized rewards, an exponential-moving-average baseline, and operation-head entropy regularization; the batch size follows the active profile in \textbf{Table~\ref{tab:hyperparams}}.  If $\hat{R}_i$ denotes the batch-normalized reward and $b$ the EMA baseline, the policy objective is \[ \max_{\theta} \frac{1}{B}\sum_{i=1}^{B}(\hat{R}_i-b)\log \pi_{\theta}(a_i\mid o_i) + \beta\,\mathcal{H}(\pi_{\theta}^{\mathrm{op}}), \] with Adam optimization and gradient clipping.  Precise leave-one-out credit is refreshed periodically once the pool has reached the minimum size specified by the active profile in \textbf{Table~\ref{tab:hyperparams}}; each refresh samples up to three training tasks and updates the historical credit state role by role.

Before an edit can be committed, the trainer also enforces the structural constraints used in the main system: pool-size bounds, protected-role removal blocking, seed-family coverage preservation, validator-count minima under the reserved validator pass, rejection of newly generated validator roles when the reserved validator pass is enabled, and diversity-based rejection for dominant-family or near-duplicate additions. During the main phase, training stops early if the committed pool stays unchanged for more than 85\% of the most recent 24 main-phase episodes after at least one full main-phase window.

\subsection{Credit-Ranked DAG Construction}
\label{app:credit_ranked_dag}

This subsection makes explicit the graph-construction routine used inside the inference operator.  Given the retrieved non-terminal set $B_t=A_t\setminus\{r_{\mathrm{agg}}\}$, the procedure first imposes the stage-aware total order from \Cref{eq:dag_order}.  Setup roles are therefore placed before specialist roles, validators are delayed until after specialist reasoning, and roles within the same stage are ordered by decreasing current fast credit with a deterministic index tie-breaker.  In the reported experiments, when degree caps are not specified externally, the routine uses $b_{\mathrm{out}}=\max(1,\min(2,n-1))$ and $b_{\mathrm{in}}=\max(1,\lfloor n/2\rfloor)$ for $n=|B_t|$.

Edges are then added only from earlier to later roles in this ordered list, subject to the out-degree budget of the source and the in-degree budget of the target.  This forward orientation makes the intermediate communication graph acyclic by construction, while the final edges from every non-terminal role to $r_{\mathrm{agg}}$ add a single terminal sink that preserves aggregator observability.  The resulting graph is finally decomposed into dependency levels: all roles in one level have no unmet predecessor in $B_t$ and can be invoked in parallel under the same upstream-message semantics used by $F$.

\begin{algorithm*}[tbh]
  \caption{Credit-ranked DAG construction for an active team}
  \label{alg:credit_ranked_dag}
\begin{algorithmic}[1]
\State \textbf{Input:} non-terminal active roles $B_t$, role types $\nu_i$, fast credits $c_i^{\mathrm{fast}}$, terminal aggregator $r_{\mathrm{agg}}$, optional caps $b_{\mathrm{in}}, b_{\mathrm{out}}$
\State \textbf{Output:} communication edge set $E_t$ and parallel levels $\mathcal{L}_t$
\Statex
\State \phaseline{// Phase 1: Stage- and credit-ranked order}
\State $n \leftarrow |B_t|$
\If{$n=0$}
  \State \Return $(\emptyset,\emptyset)$
\EndIf
\State Set $b_{\mathrm{out}}\leftarrow \max(1,\min(2,n-1))$ if unspecified
\State Set $b_{\mathrm{in}}\leftarrow \max(1,\lfloor n/2\rfloor)$ if unspecified
\ForAll{$r_i\in B_t$}
  \State $K_i\leftarrow (\kappa(\nu_i),-c_i^{\mathrm{fast}},i)$ \Comment{stage, credit, tie-breaker}
\EndFor
\State $(r_{(1)},\ldots,r_{(n)})\leftarrow$ roles in $B_t$ sorted by increasing $K_i$
\Statex
\State \phaseline{// Phase 2: Bounded forward edge construction}
\State $E_t\leftarrow\emptyset$
\ForAll{$r_i\in B_t$}
  \State $d_{\mathrm{in}}(r_i)\leftarrow 0$; $d_{\mathrm{out}}(r_i)\leftarrow 0$
\EndFor
\For{$i=1$ \textbf{to} $n$}
  \For{$j=i+1$ \textbf{to} $n$}
    \If{$d_{\mathrm{out}}(r_{(i)})\ge b_{\mathrm{out}}$}
      \State \textbf{break}
    \EndIf
    \If{$d_{\mathrm{in}}(r_{(j)})< b_{\mathrm{in}}$}
      \State $E_t\leftarrow E_t\cup\{(r_{(i)},r_{(j)})\}$
      \State $d_{\mathrm{out}}(r_{(i)})\leftarrow d_{\mathrm{out}}(r_{(i)})+1$
      \State $d_{\mathrm{in}}(r_{(j)})\leftarrow d_{\mathrm{in}}(r_{(j)})+1$
    \EndIf
  \EndFor
\EndFor
\Statex
\State \phaseline{// Phase 3: Terminal aggregation and parallel levels}
\For{$i=1$ \textbf{to} $n$}
  \State $E_t\leftarrow E_t\cup\{(r_{(i)},r_{\mathrm{agg}})\}$
\EndFor
\State $\mathcal{L}_t\leftarrow \langle\rangle$; $U\leftarrow\emptyset$; $R\leftarrow(r_{(1)},\ldots,r_{(n)})$
\While{$R\ne\emptyset$}
  \State $L\leftarrow \{r\in R:\mathrm{Pred}_{E_t}(r)\cap B_t\subseteq U\}$ preserving ranked order
  \State Append $L$ to $\mathcal{L}_t$
  \State $U\leftarrow U\cup L$; $R\leftarrow R\setminus L$
\EndWhile
\State \Return $(E_t,\mathcal{L}_t)$
\end{algorithmic}
\end{algorithm*}

\subsection{Training Loop}
\label{app:training_loop}

For reference, \Cref{alg:training_step} states one full training step end-to-end, combining the pre-edit rollout, observation construction, candidate generation, and score-gated commitment described above.

\begin{algorithm*}[tbh]
  \caption{One training step of \sero}
  \label{alg:training_step}
\begin{algorithmic}[1]
\State \textbf{Input:} task $x_t$, committed state $(\roles_t,C_t)$, phase $\chi_t$
\Statex \hspace{\algorithmicindent} inference operator $F$, encoder $e$, credit summary $\psi$, action-mask rule $\mathcal{M}$
\Statex \hspace{\algorithmicindent} controller $\pi_\theta$, candidate generator $\mathcal{E}$, commitment rule $\Gamma$, scorer $s_{x_t}$
\State \textbf{Output:} committed state $(\roles_{t+1}, C_{t+1})$ and reward $R_t$
\Statex
\State \phaseline{Phase 1: Pre-edit rollout, observation, and action sampling}
\State $(\hat y_t, A_t, C_t^{\mathrm{pre}}) \leftarrow F(x_t; \roles_t, C_t)$ \Comment{current-pool inference}
\State $o_t \leftarrow \bigl[e(x_t,\hat y_t),\,\bar e_{A_t},\,\psi(C_t^{\mathrm{pre}})\bigr]$ \Comment{observation from \Cref{eq:observation}}
\State $\mathcal{M}_t \leftarrow \mathcal{M}(\roles_t,C_t^{\mathrm{pre}},\chi_t)$ \Comment{contract-defined action mask}
\State Sample $a_t = (\omega_t, z_t) \sim \pi_\theta(\cdot \mid o_t, \mathcal{M}_t)$ \Comment{$\omega_t\!\in\!\mathcal{O}$}
\Statex
\State \phaseline{Phase 2: Candidate generation and re-scoring}
\If{$\omega_t = \textsc{Noop}$}
  \State $(\roles'_t, C'_t, \hat y'_t, R_t) \leftarrow (\roles_t, C_t^{\mathrm{pre}}, \hat y_t, 0)$
\Else
  \State $(\roles'_t, C'_t) \leftarrow \mathcal{E}(a_t,x_t; \roles_t, C_t^{\mathrm{pre}})$ \Comment{contract-checked candidate state}
  \If{$(\roles'_t,C'_t)=(\roles_t,C_t^{\mathrm{pre}})$} \Comment{candidate unchanged}
    \State $\hat y'_t \leftarrow \hat y_t$ and $R_t \leftarrow 0$
  \Else
    \State $(\hat y'_t, A'_t, C'_t) \leftarrow F(x_t; \roles'_t, C'_t)$
    \State $R_t \leftarrow s_{x_t}(\hat y'_t) - s_{x_t}(\hat y_t)$ \Comment{candidate-vs-current score change}
  \EndIf
\EndIf
\Statex
\State \phaseline{Phase 3: Policy update and score-gated commitment}
\State Update $\theta$ with REINFORCE using reward $R_t$
\If{$\Gamma\bigl((\roles_t, C_t^{\mathrm{pre}}),\,(\roles'_t, C'_t),\,R_t,\,\chi_t,\,x_t\bigr)=1$}
  \State $(\roles_{t+1}, C_{t+1}) \leftarrow (\roles'_t, C'_t)$ \Comment{commitment rule accepts candidate}
\Else
  \State $(\roles_{t+1}, C_{t+1}) \leftarrow (\roles_t, C_t^{\mathrm{pre}})$ \Comment{restore pre-edit state}
\EndIf
\end{algorithmic}
\end{algorithm*}

\section{Additional Experimental Results and Mechanism Analyses}
\label{app:additional_results}

This appendix reports the seed-level results and supplementary mechanism analyses referenced in the main experimental section. All analyses are derived from the original training and evaluation runs without additional model queries.

\subsection{Seed-Level Detailed Results}
\label{app:per_seed}

\textbf{Table~\ref{tab:app_seed_all}} reports the seed-level results underlying the main averaged table. For \naturalplan, each seed reports partial (P) and exact (E) accuracy; \olympiadbench and \tablebench report task score. All entries are percentages.  \cellcolor{bestshade}\textbf{Best} and \cellcolor{secondshade}\underline{second-best} per column within each backbone block.

\begin{table*}[tbh]
\centering
\scriptsize
\setlength{\tabcolsep}{2.6pt}
\renewcommand{\arraystretch}{1.18}
\resizebox{\textwidth}{!}{%
\begin{tabular}{@{}llcccccccccccc@{}}
\toprule
\multirow{2}{*}{Backbone} & \multirow{2}{*}{Method} & \multicolumn{6}{c}{\naturalplan} & \multicolumn{3}{c}{\olympiadbench} & \multicolumn{3}{c}{\tablebench} \\
\cmidrule(lr){3-8}\cmidrule(lr){9-11}\cmidrule(lr){12-14}
 &  & Seed 42 P & Seed 42 E & Seed 43 P & Seed 43 E & Seed 44 P & Seed 44 E & Seed 42 & Seed 43 & Seed 44 & Seed 42 & Seed 43 & Seed 44 \\
\midrule
\multirow{7}{*}{GPT-4o-mini}
 & CoT                       & 49.66 & 22.22 & 50.63 & 21.89 & 50.49 & 22.22 & 37.65 & 38.25 & 37.23 & 47.49 & 46.98 & 47.24 \\
 & SC@3                      & 50.59 & 22.78 & 50.71 & 21.89 & 50.53 & 21.78 & 39.48 & \cellcolor{bestshade}\textbf{40.76} & 38.60 & 48.24 & 46.73 & 48.99 \\
 & Static DAG MAS            & 55.07 & 20.56 & 55.60 & 20.78 & \cellcolor{secondshade}\underline{57.31} & 22.00 & 38.76 & 38.52 & 38.10 & 51.51 & \cellcolor{secondshade}\underline{51.76} & \cellcolor{secondshade}\underline{53.64} \\
 & Workflow                  & 57.12 & 25.56 & \cellcolor{secondshade}\underline{57.07} & 25.33 & 57.29 & 23.44 & 33.06 & 35.75 & 35.83 & 48.12 & 49.75 & 49.37 \\
 & Random Role Evolution     & \cellcolor{bestshade}\textbf{58.27} & \cellcolor{secondshade}\underline{28.33} & 54.92 & 25.11 & 52.12 & 22.33 & \cellcolor{bestshade}\textbf{41.54} & 39.14 & 34.95 & 51.88 & 51.26 & 52.14 \\
 & Static Role Orchestration & 55.84 & 27.33 & \cellcolor{bestshade}\textbf{57.77} & \cellcolor{secondshade}\underline{29.11} & 57.10 & \cellcolor{secondshade}\underline{28.00} & 38.99 & 38.76 & \cellcolor{secondshade}\underline{39.17} & \cellcolor{secondshade}\underline{53.89} & 51.38 & 52.01 \\
 & \sero~(Ours)              & \cellcolor{secondshade}\underline{58.02} & \cellcolor{bestshade}\textbf{30.67} & 57.06 & \cellcolor{bestshade}\textbf{29.33} & \cellcolor{bestshade}\textbf{58.39} & \cellcolor{bestshade}\textbf{31.11} & \cellcolor{secondshade}\underline{40.82} & \cellcolor{secondshade}\underline{40.55} & \cellcolor{bestshade}\textbf{39.43} & \cellcolor{bestshade}\textbf{54.52} & \cellcolor{bestshade}\textbf{53.39} & \cellcolor{bestshade}\textbf{54.27} \\
\midrule
\multirow{7}{*}{Gemini-2.5-flash-lite}
 & CoT                       & 59.54 & 30.89 & 59.18 & 31.11 & 59.30 & 30.89 & 40.81 & 40.88 & 41.11 & 58.92 & 59.30 & 58.79 \\
 & SC@3                      & 59.22 & 29.89 & 59.19 & 30.78 & 59.45 & 30.78 & 41.57 & 40.24 & 40.97 & 58.17 & 59.30 & 58.04 \\
 & Static DAG MAS            & \cellcolor{secondshade}\underline{78.09} & \cellcolor{secondshade}\underline{55.22} & \cellcolor{secondshade}\underline{78.03} & \cellcolor{secondshade}\underline{55.22} & \cellcolor{secondshade}\underline{76.91} & \cellcolor{secondshade}\underline{53.67} & 51.28 & 51.50 & 50.28 & 60.68 & 60.55 & 60.80 \\
 & Workflow                  & 72.47 & 51.44 & 72.69 & 51.22 & 72.56 & 50.78 & \cellcolor{secondshade}\underline{60.27} & \cellcolor{secondshade}\underline{60.59} & 61.35 & 57.41 & 57.91 & 58.42 \\
 & Random Role Evolution     & 62.83 & 40.78 & 56.48 & 36.67 & 70.92 & 44.67 & 59.90 & 57.12 & 50.87 & 59.42 & 61.18 & \cellcolor{secondshade}\underline{61.31} \\
 & Static Role Orchestration & 69.32 & 43.44 & 70.78 & 44.33 & 70.03 & 44.00 & 41.39 & 55.10 & \cellcolor{secondshade}\underline{62.42} & \cellcolor{secondshade}\underline{61.68} & \cellcolor{secondshade}\underline{61.81} & \cellcolor{secondshade}\underline{61.31} \\
 & \sero~(Ours)              & \cellcolor{bestshade}\textbf{80.74} & \cellcolor{bestshade}\textbf{56.44} & \cellcolor{bestshade}\textbf{80.07} & \cellcolor{bestshade}\textbf{56.56} & \cellcolor{bestshade}\textbf{81.31} & \cellcolor{bestshade}\textbf{57.33} & \cellcolor{bestshade}\textbf{64.06} & \cellcolor{bestshade}\textbf{66.02} & \cellcolor{bestshade}\textbf{65.37} & \cellcolor{bestshade}\textbf{61.81} & \cellcolor{bestshade}\textbf{63.69} & \cellcolor{bestshade}\textbf{64.95} \\
\midrule
\multirow{7}{*}{Qwen3-8b}
 & CoT                       & 45.98 & 13.67 & 46.14 & 13.78 & 46.20 & 14.00 & 35.58 & 36.24 & 34.87 & 29.40 & 29.40 & 29.52 \\
 & SC@3                      & 46.08 & 13.89 & 46.22 & 13.78 & 46.27 & 13.78 & 35.54 & 37.33 & 36.93 & 29.40 & 29.27 & 29.15 \\
 & Static DAG MAS            & \cellcolor{bestshade}\textbf{54.89} & \cellcolor{bestshade}\textbf{19.33} & \cellcolor{bestshade}\textbf{54.35} & \cellcolor{bestshade}\textbf{20.11} & \cellcolor{bestshade}\textbf{55.45} & \cellcolor{bestshade}\textbf{21.44} & 43.64 & 42.32 & 42.72 & 45.23 & 43.59 & 43.34 \\
 & Workflow                  & \cellcolor{secondshade}\underline{50.50} & \cellcolor{secondshade}\underline{15.56} & \cellcolor{secondshade}\underline{50.08} & 15.78 & \cellcolor{secondshade}\underline{50.68} & \cellcolor{secondshade}\underline{16.00} & 40.04 & 38.50 & 41.73 & 44.85 & 43.84 & 44.60 \\
 & Random Role Evolution     & 19.72 &  0.56 & 49.64 & 14.67 & 48.49 & 15.56 & \cellcolor{bestshade}\textbf{50.13} & \cellcolor{bestshade}\textbf{50.59} & 41.89 & \cellcolor{bestshade}\textbf{52.39} & 47.24 & 45.35 \\
 & Static Role Orchestration & 47.87 & 13.67 & 48.14 & 13.33 & 47.74 & 13.78 & \cellcolor{secondshade}\underline{49.28} & 46.96 & \cellcolor{secondshade}\underline{47.44} & 48.12 & \cellcolor{secondshade}\underline{48.74} & \cellcolor{secondshade}\underline{46.98} \\
 & \sero~(Ours)              & 46.39 & 15.00 & 47.05 & \cellcolor{secondshade}\underline{16.22} & 45.89 & 13.67 & 48.85 & \cellcolor{secondshade}\underline{47.82} & \cellcolor{bestshade}\textbf{49.14} & \cellcolor{secondshade}\underline{49.50} & \cellcolor{bestshade}\textbf{49.50} & \cellcolor{bestshade}\textbf{49.12} \\
\bottomrule
\end{tabular}}
\caption{Seed-level results by backbone.  P and E denote \naturalplan partial and exact accuracy.  \textbf{Best} and \underline{second-best} per column within each backbone block.}
\label{tab:app_seed_all}
\end{table*}

\subsection{Role-Pool Evolution}
\label{app:role_lifecycle_analysis}

\paragraph{Role Lifecycle Statistics.}
\textbf{Table~\ref{tab:app_role_lifecycle}} tracks role-pool evolution by model, benchmark, and seed, jointly characterizing whether evolution produces durable specialists rather than merely enlarging the pool.

\begin{table*}[tbh]
\centering
\scriptsize
\setlength{\tabcolsep}{3.2pt}
\renewcommand{\arraystretch}{1.08}
\resizebox{\textwidth}{!}{%
\begin{tabular}{@{}llrrrrlll@{}}
\toprule
Model & Bench. & Seed & Added & Removed & Surv. additions & Survival rate & Evolved roles used & Unused-role ratio \\
\midrule
Gemini & NP & 42 & 0 & 0 & 0 & -- & -- & 40.0\% \\
Gemini & NP & 43 & 0 & 0 & 0 & -- & -- & 30.0\% \\
Gemini & NP & 44 & 0 & 0 & 0 & -- & -- & 40.0\% \\
Gemini & OB & 42 & 4 & 1 & 3 & 75.0\% & 3/3 (100.0\%) & 10.0\% \\
Gemini & OB & 43 & 3 & 0 & 3 & 100.0\% & 2/3 (66.7\%) & 20.0\% \\
Gemini & OB & 44 & 3 & 0 & 3 & 100.0\% & 3/3 (100.0\%) & 0.0\% \\
Gemini & TB & 42 & 3 & 0 & 3 & 100.0\% & 2/3 (66.7\%) & 30.0\% \\
Gemini & TB & 43 & 3 & 0 & 3 & 100.0\% & 3/3 (100.0\%) & 10.0\% \\
Gemini & TB & 44 & 3 & 1 & 2 & 66.7\% & 2/2 (100.0\%) & 11.1\% \\
GPT & NP & 42 & 0 & 0 & 0 & -- & -- & 0.0\% \\
GPT & NP & 43 & 0 & 0 & 0 & -- & -- & 30.0\% \\
GPT & NP & 44 & 0 & 0 & 0 & -- & -- & 20.0\% \\
GPT & OB & 42 & 3 & 0 & 3 & 100.0\% & 3/3 (100.0\%) & 10.0\% \\
GPT & OB & 43 & 3 & 1 & 2 & 66.7\% & 2/2 (100.0\%) & 22.2\% \\
GPT & OB & 44 & 3 & 0 & 3 & 100.0\% & 2/3 (66.7\%) & 20.0\% \\
GPT & TB & 42 & 4 & 2 & 2 & 50.0\% & 0/2 (0.0\%) & 22.2\% \\
GPT & TB & 43 & 5 & 3 & 2 & 40.0\% & 2/2 (100.0\%) & 11.1\% \\
GPT & TB & 44 & 3 & 0 & 3 & 100.0\% & 1/3 (33.3\%) & 20.0\% \\
Qwen & NP & 42 & 2 & 1 & 2 & 100.0\% & 2/2 (100.0\%) & 33.3\% \\
Qwen & NP & 43 & 3 & 1 & 2 & 66.7\% & 1/2 (50.0\%) & 41.7\% \\
Qwen & NP & 44 & 2 & 2 & 0 & 0.0\% & -- & 60.0\% \\
Qwen & OB & 42 & 6 & 2 & 4 & 66.7\% & 4/4 (100.0\%) & 0.0\% \\
Qwen & OB & 43 & 6 & 1 & 5 & 83.3\% & 5/5 (100.0\%) & 0.0\% \\
Qwen & OB & 44 & 5 & 0 & 5 & 100.0\% & 5/5 (100.0\%) & 8.3\% \\
Qwen & TB & 42 & 6 & 1 & 5 & 83.3\% & 0/5 (0.0\%) & 50.0\% \\
Qwen & TB & 43 & 5 & 1 & 4 & 80.0\% & 0/4 (0.0\%) & 45.5\% \\
Qwen & TB & 44 & 5 & 0 & 5 & 100.0\% & 0/5 (0.0\%) & 50.0\% \\
\bottomrule
\end{tabular}}
\caption{Role lifecycle statistics.  Gemini denotes Gemini-2.5-flash-lite, GPT denotes GPT-4o-mini, Qwen denotes Qwen3-8b, and NP/OB/TB denote \naturalplan, \olympiadbench, and \tablebench.}
\label{tab:app_role_lifecycle}
\end{table*}

The pattern is benchmark- and backbone-dependent. \naturalplan keeps the seed pool nearly intact under Gemini and GPT-4o-mini, \olympiadbench retains most added roles for downstream use, and \tablebench is mixed with some additions going unused at evaluation time. Role evolution therefore behaves as a selective mechanism whose intensity is benchmark- and backbone-conditioned rather than uniformly active.

\begin{figure*}[tbh]
\centering
\includegraphics[width=\textwidth]{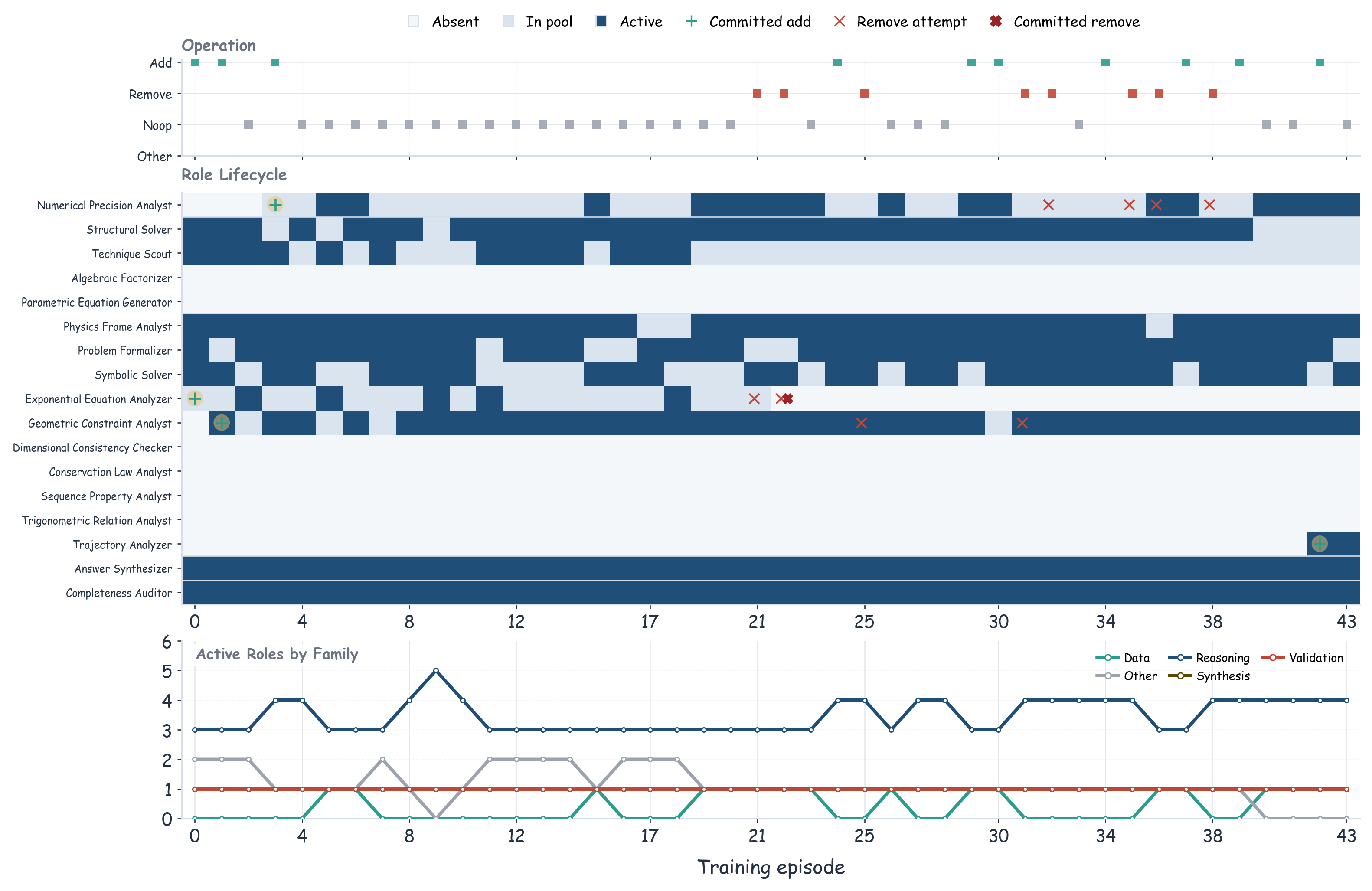}
\caption{Role lifecycle of a \sero trajectory on \olympiadbench with Gemini-2.5-flash-lite (seed 42).  \textbf{Top}: controller operation per training episode, with $\checkmark$ and $\times$ marking committed and rejected edits.  \textbf{Middle}: lifecycle state of every persistent role across episodes, colored as absent, in pool, or active.  \textbf{Bottom}: family-level composition of active roles per episode.}
\label{fig:app_role_lifecycle}
\end{figure*}

\paragraph{Case Study of Role Lifecycle Over Training.}
\textbf{Fig.~\ref{fig:app_role_lifecycle}} visualizes a representative \sero trajectory for Gemini-2.5-flash-lite on \olympiadbench, tracing structural interventions, role lifecycle states, and the family-level composition of active roles over training. The figure complements \textbf{Table~\ref{tab:app_role_lifecycle}} by showing when pool revisions occur and whether the revised roles subsequently enter the inference topology.

\subsection{Inference-Time Orchestration}
\label{app:active_set_diversity}

\paragraph{Evaluation Active-Set Diversity.}
\textbf{Table~\ref{tab:app_active_set_diversity}} measures evaluation-time routing diversity through the number of distinct active sets, their entropy and Simpson concentration, the mean active-role count, and the unused-role ratio.

\begin{table*}[tbh]
\centering
\scriptsize
\setlength{\tabcolsep}{3pt}
\renewcommand{\arraystretch}{1.08}
\resizebox{\textwidth}{!}{%
\begin{tabular}{@{}llrrrrrrr@{}}
\toprule
Model & Bench. & Seed & Unique active sets & Unique ratio & Entropy & Simpson & Mean active roles & Unused-role ratio \\
\midrule
Gemini & NP & 42 & 9 & 0.010 & 0.674 & 0.727 & 6.000 & 0.400 \\
Gemini & NP & 43 & 9 & 0.010 & 0.674 & 0.727 & 6.000 & 0.300 \\
Gemini & NP & 44 & 5 & 0.006 & 0.807 & 0.705 & 6.000 & 0.400 \\
Gemini & OB & 42 & 244 & 0.278 & 0.908 & 0.989 & 7.000 & 0.100 \\
Gemini & OB & 43 & 13 & 0.015 & 0.779 & 0.832 & 6.992 & 0.200 \\
Gemini & OB & 44 & 303 & 0.345 & 0.893 & 0.989 & 6.992 & 0.000 \\
Gemini & TB & 42 & 3 & 0.004 & 0.639 & 0.496 & 7.000 & 0.300 \\
Gemini & TB & 43 & 13 & 0.016 & 0.547 & 0.660 & 7.000 & 0.100 \\
Gemini & TB & 44 & 8 & 0.010 & 0.551 & 0.610 & 7.000 & 0.111 \\
GPT & NP & 42 & 15 & 0.017 & 0.743 & 0.843 & 6.000 & 0.000 \\
GPT & NP & 43 & 3 & 0.003 & 1.000 & 0.667 & 6.000 & 0.300 \\
GPT & NP & 44 & 7 & 0.008 & 0.732 & 0.726 & 6.000 & 0.200 \\
GPT & OB & 42 & 469 & 0.535 & 0.948 & 0.995 & 6.960 & 0.100 \\
GPT & OB & 43 & 12 & 0.014 & 0.667 & 0.754 & 6.968 & 0.222 \\
GPT & OB & 44 & 246 & 0.281 & 0.895 & 0.988 & 6.984 & 0.200 \\
GPT & TB & 42 & 22 & 0.028 & 0.716 & 0.848 & 7.000 & 0.222 \\
GPT & TB & 43 & 28 & 0.035 & 0.781 & 0.891 & 7.000 & 0.111 \\
GPT & TB & 44 & 29 & 0.036 & 0.672 & 0.848 & 7.000 & 0.200 \\
Qwen & NP & 42 & 6 & 0.007 & 0.733 & 0.700 & 3.996 & 0.333 \\
Qwen & NP & 43 & 4 & 0.004 & 0.583 & 0.516 & 4.000 & 0.417 \\
Qwen & NP & 44 & 1 & 0.001 & 0.000 & 0.000 & 4.000 & 0.600 \\
Qwen & OB & 42 & 51 & 0.058 & 0.801 & 0.927 & 5.000 & 0.000 \\
Qwen & OB & 43 & 9 & 0.010 & 0.721 & 0.752 & 4.994 & 0.000 \\
Qwen & OB & 44 & 172 & 0.196 & 0.876 & 0.981 & 4.977 & 0.083 \\
Qwen & TB & 42 & 19 & 0.024 & 0.729 & 0.844 & 4.987 & 0.500 \\
Qwen & TB & 43 & 19 & 0.024 & 0.729 & 0.844 & 4.987 & 0.455 \\
Qwen & TB & 44 & 19 & 0.024 & 0.729 & 0.844 & 4.987 & 0.500 \\
\bottomrule
\end{tabular}}
\caption{Evaluation-time active-set diversity.  Model and benchmark abbreviations follow \textbf{Table~\ref{tab:app_role_lifecycle}}.}
\label{tab:app_active_set_diversity}
\end{table*}

Routing diversity is strongest on \olympiadbench (hundreds of distinct active sets, near-maximal Simpson scores) and weakest on \naturalplan, consistent with the latter's stricter output requirements and more repetitive subtask structure. \tablebench is intermediate but backbone-dependent, with GPT-4o-mini showing more active-set diversity than Gemini-2.5-flash-lite and Qwen3-8b cycling through a small but nontrivial family of configurations. The pattern is more consistent with task-conditioned routing than with stochastic variation alone.

\begin{figure*}[tbh]
\centering
\includegraphics[width=0.96\textwidth]{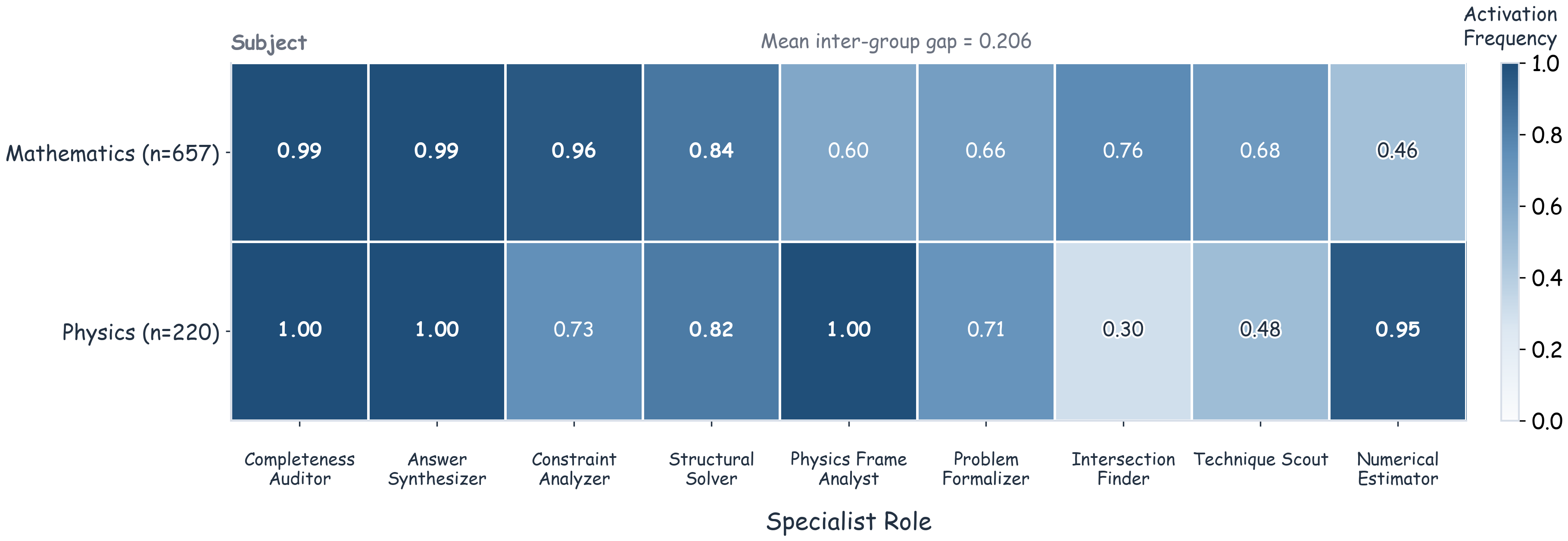}
\caption{Task-conditioned specialist-role activation on \olympiadbench with GPT-4o-mini (seed 42).  Rows are subject groups, columns are specialist roles, and each cell gives the fraction of in-group instances that activate the role.  Mean inter-group gap is $0.206$.}
\label{fig:app_task_conditioned_gpt}
\end{figure*}

\begin{figure*}[tbh]
\centering
\includegraphics[width=0.96\textwidth]{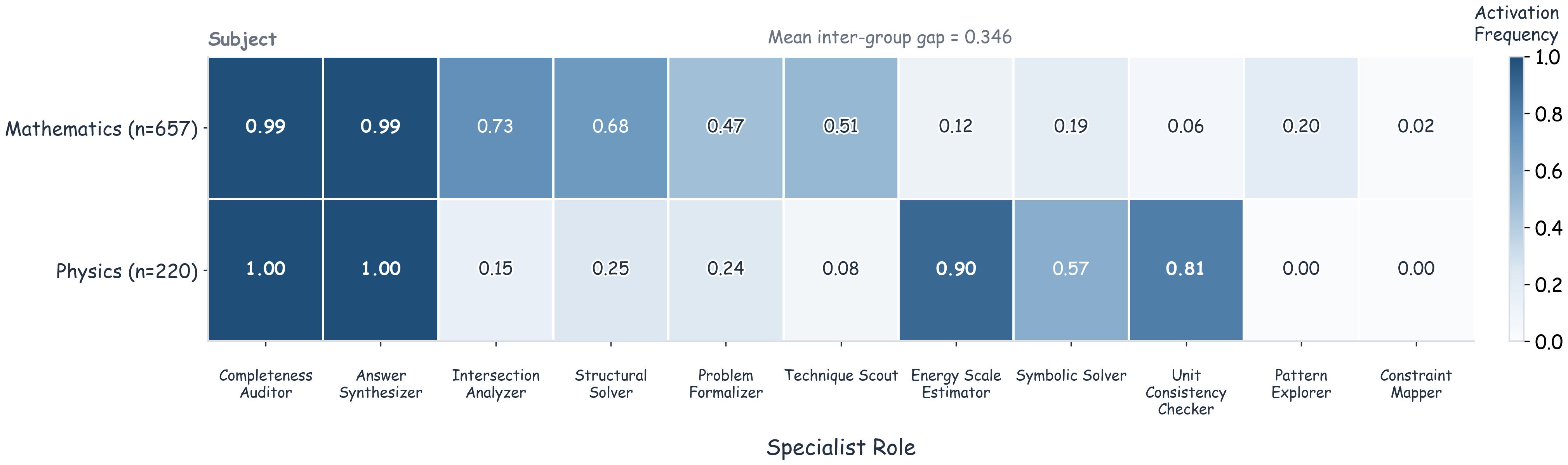}
\caption{Task-conditioned specialist-role activation on \olympiadbench with Qwen3-8B (seed 44), in the same format as \textbf{Fig.~\ref{fig:app_task_conditioned_gpt}}.  Mean inter-group gap is $0.346$.}
\label{fig:app_task_conditioned_qwen}
\end{figure*}

\paragraph{Task-Conditioned Role Activation.}
\textbf{Figs.~\ref{fig:app_task_conditioned_gpt} and \ref{fig:app_task_conditioned_qwen}} test whether active-role variation is task-conditioned. Rows are subject-level task groups, columns are interpretable specialist roles, and each cell is the fraction of instances in a group that activate the role. The question is whether routing collapses to a universal role set or varies with task structure.

In both backbones, specialist usage concentrates by subject group rather than spreading uniformly, arguing against a universal role set or noisy switching. GPT-4o-mini shows richer subject-level differentiation, consistent with its larger number of unique active sets in \textbf{Table~\ref{tab:app_active_set_diversity}}, while Qwen3-8b retains task-conditioned specialization despite a smaller active team. The gains on \olympiadbench and \tablebench therefore plausibly arise from selecting a better-matched specialist subset rather than from activating more roles.

The signature-level companion appears in the main text (\textbf{Fig.~\ref{fig:active_set_signatures}}). Together with the heatmaps, the picture is one of reusable yet conditionally adaptive routing rather than per-instance idiosyncratic graphs, clearest on \olympiadbench where adaptive routing yields the strongest gains in the main results.

\subsection{Credit and Topology Alignment}
\label{app:credit_cost_analysis}

\paragraph{Credit / DAG Alignment.}
\textbf{Table~\ref{tab:app_credit_alignment}} tests whether \sero's credit and topology mechanisms leave measurable signatures in the learned pool through three statistics, the correlation between credit and evaluation-time selection frequency, the pre-credit of removed roles, and the average credit of early versus late DAG positions.

\begin{table*}[tbh]
\centering
\scriptsize
\setlength{\tabcolsep}{4pt}
\renewcommand{\arraystretch}{1.08}
\resizebox{\textwidth}{!}{%
\begin{tabular}{@{}llrrrrr@{}}
\toprule
Model & Bench. & Seed & Credit-usage corr. & Removed pre-credit & Early-DAG credit & Late-DAG credit \\
\midrule
Gemini & NP & 42 & 0.828 & -- & 0.945 & 0.908 \\
Gemini & NP & 43 & 0.495 & -- & 0.941 & 0.864 \\
Gemini & NP & 44 & 0.878 & -- & 0.942 & 0.929 \\
Gemini & OB & 42 & -0.324 & -- & 0.947 & 0.936 \\
Gemini & OB & 43 & -0.158 & -- & 0.948 & 0.937 \\
Gemini & OB & 44 & 0.000 & -- & 0.940 & 0.916 \\
Gemini & TB & 42 & -- & -- & 0.898 & 0.896 \\
Gemini & TB & 43 & -0.612 & -- & 0.895 & 0.891 \\
Gemini & TB & 44 & 0.000 & -- & 0.857 & 0.567 \\
GPT & NP & 42 & -0.455 & -- & 0.933 & 0.819 \\
GPT & NP & 43 & -0.131 & -- & 0.940 & 0.821 \\
GPT & NP & 44 & 0.223 & -- & 0.930 & 0.841 \\
GPT & OB & 42 & -0.324 & -- & 0.954 & 0.933 \\
GPT & OB & 43 & -- & -- & 0.950 & 0.921 \\
GPT & OB & 44 & 0.126 & -- & 0.944 & 0.926 \\
GPT & TB & 42 & -- & -- & 0.885 & 0.890 \\
GPT & TB & 43 & 0.412 & 0.907 & 0.891 & 0.862 \\
GPT & TB & 44 & -0.408 & -- & 0.886 & 0.880 \\
Qwen & NP & 42 & 0.000 & 0.903 & 0.933 & 0.805 \\
Qwen & NP & 43 & 0.000 & 0.903 & 0.932 & 0.785 \\
Qwen & NP & 44 & -- & 0.890 & 0.875 & 0.857 \\
Qwen & OB & 42 & -0.335 & 0.936 & 0.951 & 0.902 \\
Qwen & OB & 43 & 0.000 & -- & 0.939 & 0.909 \\
Qwen & OB & 44 & -0.872 & -- & 0.952 & 0.938 \\
Qwen & TB & 42 & -0.410 & -- & 0.860 & 0.859 \\
Qwen & TB & 43 & -0.410 & -- & 0.860 & 0.859 \\
Qwen & TB & 44 & -0.410 & -- & 0.860 & 0.859 \\
\bottomrule
\end{tabular}}
\caption{Credit and DAG alignment statistics.  Model and benchmark abbreviations follow \textbf{Table~\ref{tab:app_role_lifecycle}}.}
\label{tab:app_credit_alignment}
\end{table*}

Early-DAG credit usually exceeds late-DAG credit, consistent with the intended credit-ranked ordering of the inference graph. Credit-usage correlation is mixed in sign, which we interpret cautiously since roles can receive high credit through occasional high-leverage use without being most-frequently selected. The non-trivial pre-credit values observed for some removed roles indicate that removal reflects redundancy and pool restructuring rather than elimination of the globally weakest role.

Together, these analyses provide convergent evidence about \sero's behavior. Pools evolve through persistent, reusable revisions, evaluation routing remains task-conditioned rather than collapsing to a single configuration, and learned credit aligns with the intended inference ordering.

\subsection{Scaling Behavior}
\label{app:scaling}

\textbf{Tables~\ref{tab:scaling_model} and \ref{tab:scaling_controller}} list the exact per-benchmark scores behind the curves in \textbf{Fig.~\ref{fig:scaling}}. The base-model sweep varies the Qwen3 backbone size, and the controller-width sweep varies the controller hidden width $d_h$ on Gemini-2.5-flash-lite, holding all other \sero settings fixed.  NP, OB, and TB denote \naturalplan, \olympiadbench, and \tablebench.

\begin{table*}[!t]
\centering
\footnotesize
\begin{minipage}[t]{0.45\textwidth}
\centering
\resizebox{\linewidth}{!}{%
\begin{tabular}{lcccc}
\toprule
Model & NP (P) & NP (E) & OB & TB \\
\midrule
Qwen3-1.7B & 22.62 & 3.56 & 33.88 & 23.74 \\
Qwen3-4B & 37.10 & 5.89 & 50.61 & 35.80 \\
Qwen3-8B & 48.99 & 15.22 & 52.79 & 33.04 \\
Qwen3-14B & 59.92 & 30.00 & 51.08 & 49.12 \\
Qwen3-32B & 64.69 & 39.00 & 57.99 & 55.15 \\
\bottomrule
\end{tabular}}
\caption{Base-model scaling on Qwen3 behind \textbf{Fig.~\ref{fig:scaling}}(a).  Values are mean task accuracy (\%); NP reports partial (P) and exact (E) accuracy.}
\label{tab:scaling_model}
\end{minipage}
\hfill
\begin{minipage}[t]{0.51\textwidth}
\centering

\resizebox{\linewidth}{!}{%
\begin{tabular}{rrcccc}
\toprule
$d_h$ & Params & NP (P) & NP (E) & OB & TB \\
\midrule
64 & 115{,}844 & 79.92 & 56.11 & 63.06 & 62.94 \\
128 & 256{,}068 & 79.94 & 56.11 & 64.19 & 62.31 \\
256 & 610{,}244 & 76.90 & 48.78 & 65.37 & 64.95 \\
512 & 1{,}613{,}508 & 80.01 & 55.11 & 65.06 & 62.31 \\
1024 & 4{,}799{,}684 & 79.00 & 52.78 & 63.35 & 62.44 \\
\bottomrule
\end{tabular}}
\caption{Controller-width scaling on Gemini-2.5-flash-lite behind \textbf{Fig.~\ref{fig:scaling}}(b). Values are mean task accuracy (\%); $d_h{=}256$ is the base controller.}
\label{tab:scaling_controller}
\end{minipage}
\end{table*}

\end{document}